\documentclass[sigconf]{acmart}
\usepackage{booktabs} 
\usepackage{array}  
\usepackage{graphicx} 
\usepackage{caption}
\usepackage{float} 
\usepackage{multirow}
\usepackage{graphicx}
\usepackage{tabularx}
\usepackage{xcolor}
\usepackage{colortbl}
\usepackage{siunitx}
\usepackage{graphicx}
\usepackage{amsmath}
\usepackage{algorithm}
\usepackage{algorithmic} % 或者使用 \usepackage{algpseudocode}，取决于你习惯的语法
\usepackage{multicol}
\usepackage{enumitem}

\begin{document}

\title{GeoMind: An Agentic Workflow for Lithology Classification \\with Reasoned Tool Invocation}

\author{Mingyue Cheng$^{1}$, Yitong Zhou$^{1}$, Jiahao Wang$^{1}$, Qingyang Mao$^{1}$, Qi Liu$^{1}$}

\affiliation{%
  \institution{$^1$State Key Laboratory of Cognitive Intelligence, University of Science and Technology of China, Hefei, China}
  \country{}
}
\email{{yitong.zhou, jiahaowang, maoqy0503}@mail.ustc.edu.cn}
\email{{mycheng, qiliuql}@ustc.edu.cn}

\keywords{Lithology classification, Agentic workflow, Large language models}

\begin{abstract}
\vspace{-0.04in}
Lithology classification in well logs is a fundamental geoscience data mining task that aims to infer rock types from multi dimensional geophysical sequences. Despite recent progress, existing approaches typically formulate the problem as a static, single-step discriminative mapping.
This static paradigm limits evidence-based diagnostic reasoning against geological standards, often yielding predictions that are detached from geological reality due to a lack of domain priors.
In this work, we propose GeoMind, a tool-augmented agentic framework that models lithology classification as a sequential reasoning process. GeoMind organizes its toolkit into perception, reasoning, and analysis modules, which respectively translate raw logs into semantic trends, infer lithology hypotheses from multi-source evidence, and verify predictions against stratigraphic constraints. A global planner adaptively coordinates these modules based on input characteristics, enabling geologically plausible and evidence-grounded decisions.
To guarantee the logical consistency of GeoMind, we introduce a fine-grained process supervision strategy. Unlike standard methods that focus solely on final outcomes, our approach optimizes intermediate reasoning steps, ensuring the validity of decision trajectories and alignment to geological constraints.
Experiments on four benchmark well-log datasets demonstrate that GeoMind consistently outperforms strong baselines in classification performance while providing transparent and traceable decision-making processes \footnote{The code is at \url{https://github.com/lqzxt/GeoMind}.}.
\vspace{-0.04in}
\end{abstract}
\vspace{-0.1in}

\maketitle
% {\let\thefootnote\relax\footnotetext{*Corresponding author}}

% \input{section/1-introduction}
% \input{section/1-introduction-kdd}
\begin{figure}[t] \centering
    \includegraphics[width=0.4\textwidth]{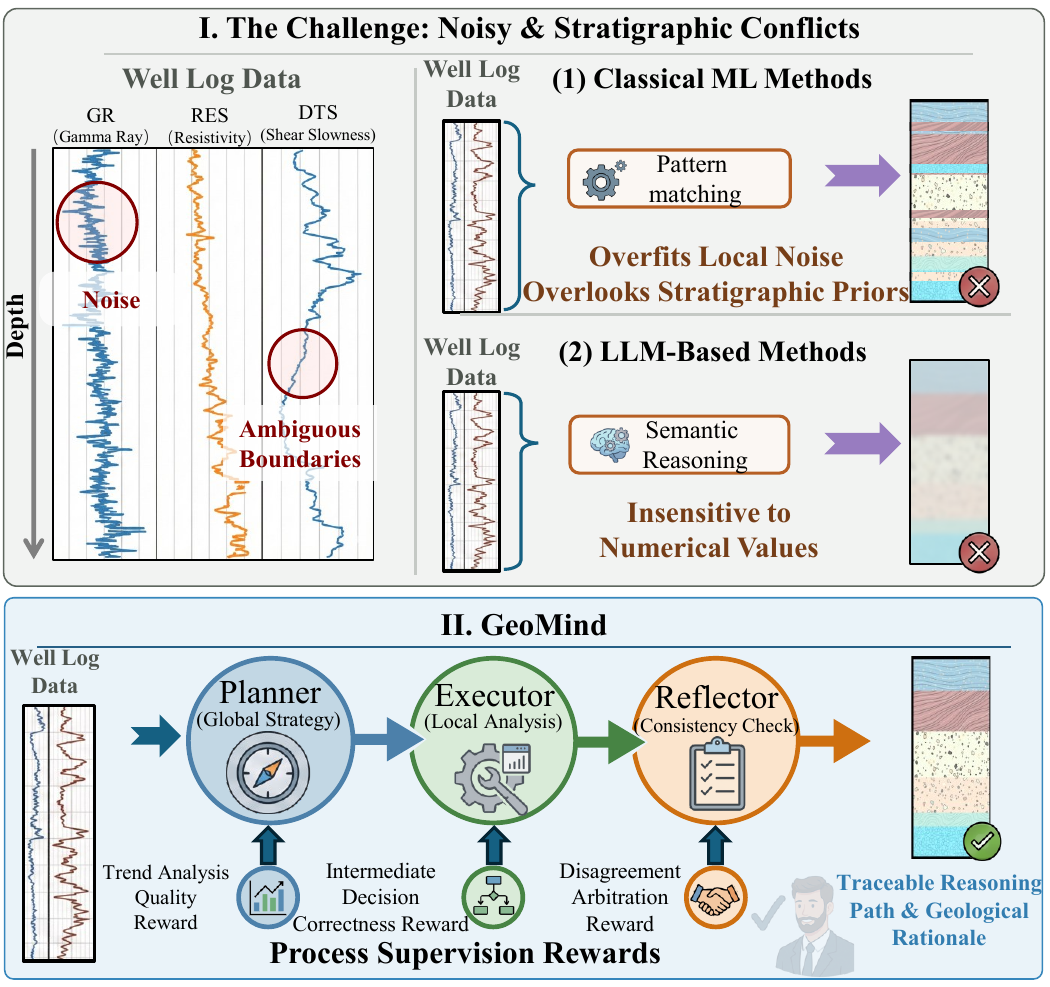}
    \vspace{-0.1in}
    \caption{Motivation of GeoMind. GeoMind addresses noisy, stratigraphically conflicting well logs through a process-supervised agentic workflow with process rewards.} \label{fig:motivation}
    \vspace{-0.2in}
\end{figure}

\vspace{-0.1in}
\section{Introduction}
Lithology classification from well-log data is a core task in subsurface characterization, supporting critical downstream applications such as reservoir evaluation, mineral exploration, and geological modeling~\cite{darling2005well}. Well logs provide dense, multivariate measurements along depth and reflect complex interactions among lithology, petrophysical properties, and measurement conditions~\cite{li2021application, doveton1985log}. From a data mining perspective, this task represents a complex sequence labeling problem where multi-dimensional, noisy measurements from well logs must be mapped to discrete geological classes along a depth axis. Mapping these numerical signals to lithological labels is challenging due to ambiguous formation boundaries and the variable responsiveness of logging tools to geological changes~\cite{el2025artificial,kim2024cafo}.

Significant advancements in this field have been driven by numerical data-driven models~\cite{bestagini2017machine,lu2025timecapsule,liu2022integrating} and the recent language-model-based methods~\cite{deng2024k2,zhou2023one}. Traditional machine learning and deep time-series models learn discriminative mappings from raw geophysical measurements to lithology labels, often performing well in controlled settings. However, they typically rely mainly on statistical associations. Consequently, they can be fragile under distribution shifts or near ambiguous boundaries, and they struggle to model the consistency constraints imposed by stratigraphic sequences~\cite{liu2021lithological}. Conversely, large language models (LLMs) are actively being explored for geological reasoning~\cite{cheng2025can,lin2023geogalactica,tao2026memcast, luo2025time}, leveraging knowledge implicitly encoded in their pretrained parameters. While LLM-based methods improve semantic grounding, they often face difficulties with numerical precision and noisy multichannel signals~\cite{fons2024evaluating, cheng2026position}. These complementary failure modes are illustrated in Figure~\ref{fig:motivation}: numerical models tend to overfit local noise and violate stratigraphic consistency, whereas LLM-based approaches struggle with fine-grained numerical patterns. 
% To combine the strengths of these paradigms, recent studies~\cite{wang2025tabletime,liu2025ts,zhang2025alphacast} have proposed hybrid frameworks that couple numerical predictors with LLM-based reasoning. 

Despite promising results, a fundamental limitation persists: most existing approaches treat lithology classification as a static, single-step mapping. In this conventional view, models infer labels in a single forward pass, effectively isolating the process from domain-specific logical constraints. Consequently, these systems lack the agency to leverage external diagnostic tools to verify their predictions. Furthermore, they are incapable of addressing intermediate reasoning errors; unlike human experts who iteratively gather evidence, static models lack the mechanism to reference neighboring context or perform self-correction based on geological plausibility~\cite{lin2025step,lightman2023let,wen2020fast, lai2025llmlight}. This structural deficiency often results in unstable predictions and opaque decision-making processes.

% Motivated by this, we propose GeoMind, an agentic framework that reframes lithology classification as a sequential multi-step reasoning. GeoMind is built upon a hierarchical toolkit organized into perception, reasoning, and analysis modules, designed to bridge the gap between numerical well-log signals and geological logic. Specifically, the workflow begins with perception components that translate raw series data into semantic trend narratives. These narratives are synthesized by reasoning engines to formulate initial hypotheses, which are audited by analytical validators against stratigraphic constraints to ensure geological plausibility. This entire process is coordinated by a global planner that adapts execution paths based on input signal characteristics. To align this complex behavior with expert knowledge, we introduce a process-supervised training strategy employing module-aware group relative policy optimization. By optimizing intermediate reasoning steps alongside final predictions, GeoMind achieves lithology classifications that are not only accurate but also supported by transparent, traceable decision paths.

These limitations suggest three design requirements for hybrid lithology interpretation. First, evidence acquisition should be adaptive, determining when numerical retrieval, trend analysis, or stratigraphic validation is needed. Second, evidence arbitration should resolve disagreements among heterogeneous predictors by considering confidence signals and geological constraints rather than fixed voting. Third, learning should align with the interpretation process by separately supervising trend interpretation, candidate assessment, and final correction. Motivated by this, we propose GeoMind, a process-supervised agentic workflow for verifiable well-log lithology classification. Its staged reasoning framework coordinates adaptive tool invocation, evidence generation, and constraint-aware conflict resolution throughout the interpretation process. Domain-adaptive supervised fine-tuning equips the LLM with lithology-specific knowledge, while process-level rewards guide intermediate reasoning and decision-making, enabling accurate and traceable classification under uncertainty. Our contributions are as follows:
\begin{itemize}
  \setlength{\itemsep}{1pt} 
  \setlength{\parskip}{2pt} 
  \item We propose GeoMind, a tool-augmented agentic workflow for lithology classification that supports adaptive tool orchestration, multi-source evidence integration, and stratigraphy-informed prediction revision.

  \item We propose a process-supervised framework that combines domain-adaptive fine-tuning with module-wise policy optimization, using process rewards to supervise intermediate reasoning beyond outcome-only feedback.
  
  \item Experiments on four public well-log benchmarks demonstrate consistent classification improvements under the evaluated settings, while producing inspectable decision traces grounded in numerical and stratigraphic evidence.
\end{itemize}

\vspace{-0.08in}
\section{Related Work}

\subsection{Lithology Classification}
\begin{figure}[t] \centering
    \includegraphics[width=0.43\textwidth]{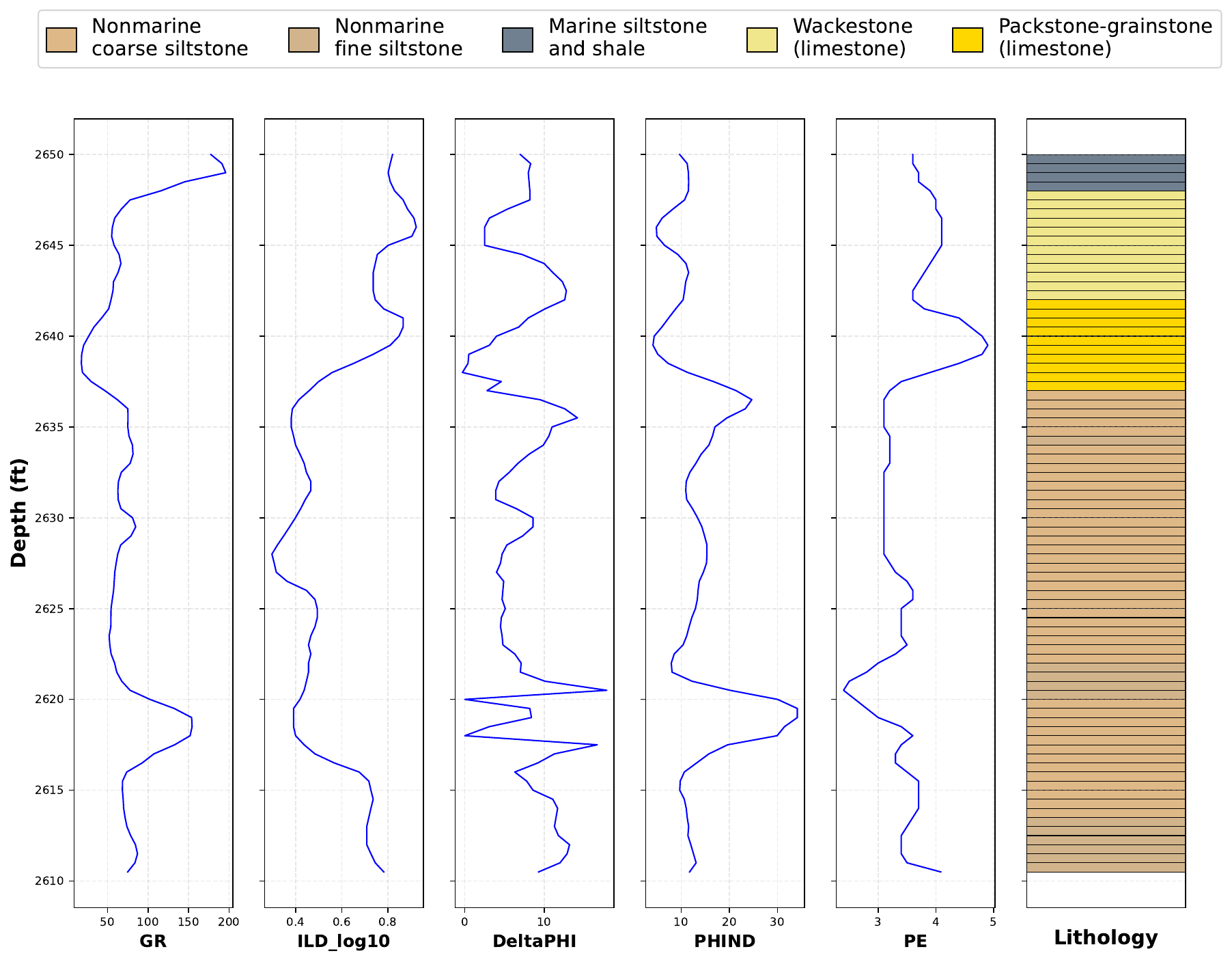}
    \vspace{-0.08in}
    \caption{Well-log curves from the facies dataset across depth for a representative interval. The data contain strong depth-wise sequential structure and lithology-grounded semantics.} \label{fig:example}
    \vspace{-0.2in}
\end{figure}

Lithology classification from well logs is a core sequence-labeling task in subsurface characterization. It provides key inputs for reservoir evaluation, stratigraphic interpretation, and geological modeling~\cite{darling2005well,doveton1985log}.
As illustrated in Figure \ref{fig:example}, well-log data exhibit strong depth-wise sequential structure and lithology-related semantics, making it challenging to model long-range stratigraphic dependencies and to disambiguate formation boundaries.
Early work typically relied on petrophysical heuristics and rule-based log analysis~\cite{asquith1982basic,doveton1985log}. Later studies adopted supervised learning over multivariate logging measurements, using classical models such as GBDT~\cite{ke2017lightgbm} and XGBoost~\cite{chen2015xgboost}.
More recent studies have shifted from handcrafted or shallow feature mapping toward deep sequence modeling, where neural architectures are used to learn local log patterns, depth-wise dependencies, and lithology transitions from multivariate well-log sequences~\cite{tschannen2017facies,xie2024transformer,li2026giat}.
To reduce geologically implausible rapid switches, some methods further incorporate stratigraphic priors through smoothing, Markov-style checks, or explicit transition constraints~\cite{liu2021lithological,schumann2002hidden}.
However, most existing approaches rely on single-stage numerical mapping or fixed structural priors, with limited integration of global stratigraphic context and cross-scale geological evidence.
This limits their interpretability and may weaken robustness in complex geological settings.

\vspace{-0.04in}
\subsection{Time Series Classification} As the well logs are a series of sensor data indicating chemical and physical characteristics of geology, the lithology classification task shares significant similarities with general time series classification. Early deep learning approaches~\cite{ismail2019deep} adapted standard neural architectures to sequential data, learning discriminative representations from local windows. To better capture temporal dependencies and multi-scale patterns, subsequent works introduced hybrid architectures such as LSTM-FCN~\cite{karim2019multivariate} and InceptionTime~\cite{ismail2020inceptiontime}. More recently, the field has witnessed a paradigm shift towards foundation models and Large Language Models. Approaches such as MOMENT~\cite{goswami2024moment} leverage pre-trained transformer backbones for universal time-series representations, while methods like GPT4TS~\cite{zhou2023one} and TableTime~\cite{wang2025tabletime} attempt to reprogram or fine-tune LLMs to process numerical sequences directly. Despite their success in general domains, these purely data-driven or implicitly knowledgeable models often struggle to handle the specific "salt-and-pepper" noise inherent in well logs or to explicitly enforce stratigraphic consistency, highlighting the need for a framework that can integrate rigorous numerical processing with verifiable geological reasoning.

\vspace{-0.04in}
\subsection{Reinforcement Learning for LLM Agents} 
Reinforcement learning has become a foundational paradigm for aligning large language models with increasingly complex constraints and reasoning tasks \cite{ouyang2022training, stiennon2020learning}. While early approaches predominantly utilized PPO \cite{schulman2017proximal}, they often suffer from training instability and high computational overhead, particularly when facing the credit assignment problem inherent in sparse, outcome-based reward settings \cite{lightman2023let}. To address this, recent advancements have shifted towards critic-free or group-relative optimization strategies, such as DAPO \cite{yu2025dapo} and GRPO \cite{shao2024deepseekmath}, which estimate baselines via group statistics to stabilize learning without a separate value network. However, when seamlessly transplanting these methods to agentic workflows~\cite{golubev2025training,jiang2025verltool}, current practices still largely rely on coarse-grained, whole-trajectory evaluations. This outcome-centric supervision inadvertently ignores the dense process signals within intermediate steps, failing to explicitly distinguish the specific contributions of planning, execution, and reflection modules.

\begin{figure*}[thp] 
    \centering
    \includegraphics[width=0.98\textwidth]{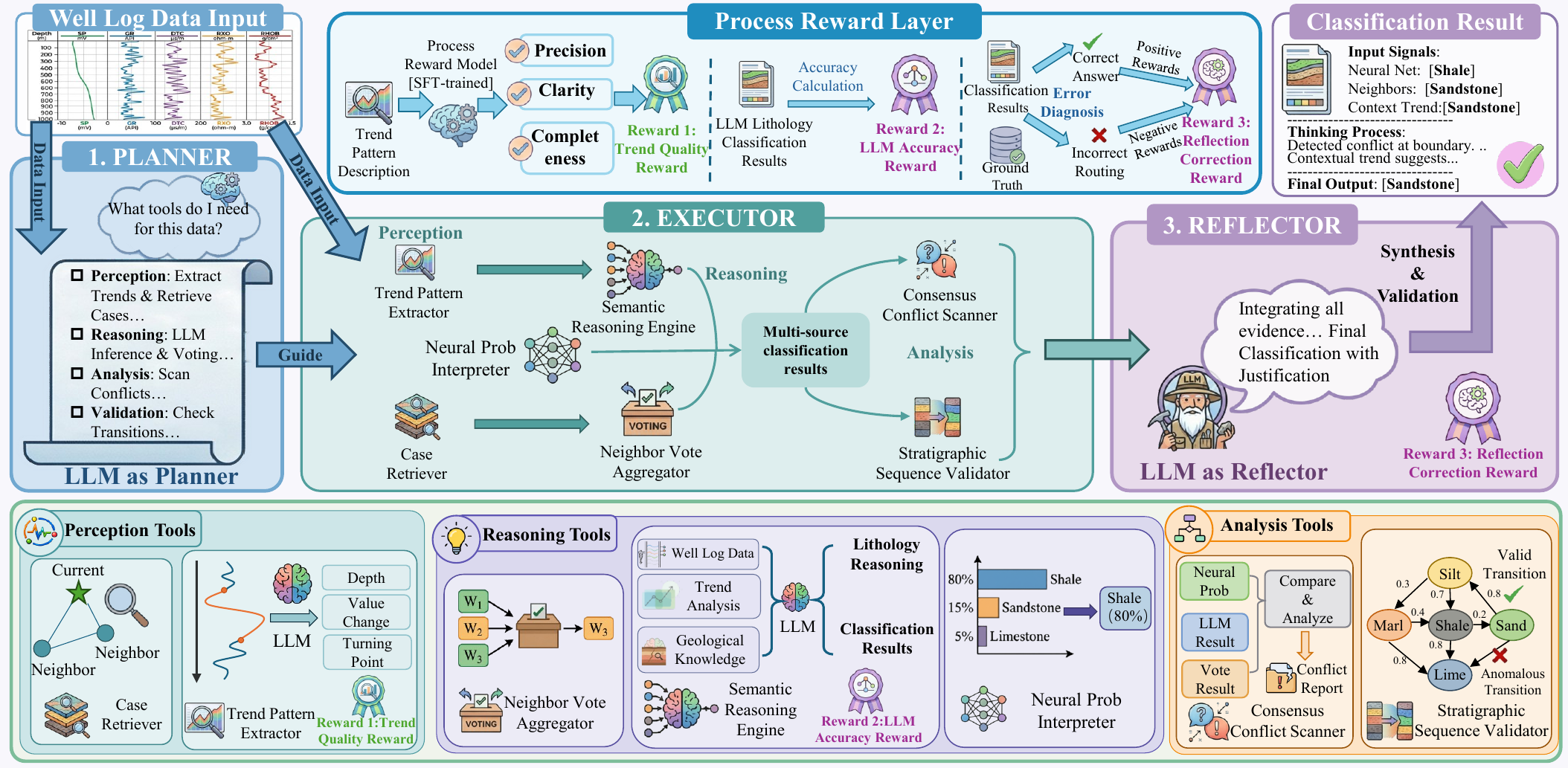} 
    \vspace{-0.1in}
    \caption{Overview of GeoMind for well-log lithology classification.
GeoMind uses a Planner–Executor–Reflector workflow with process rewards to coordinate predictors and LLM reasoning for accurate, interpretable predictions. The depicted executor workflow is shown for illustrative purposes; actual execution paths are dynamically adjusted based on the planner’s outputs.} 
    \label{fig:main_fig} 
    \vspace{-0.02in}
\end{figure*}

\section{The Proposed GeoMind}

\vspace{-0.02in}
\subsection{Problem Definition}

Let $\mathcal{D}=\{(X^{(i)},y^{(i)})\}_{i=1}^{n}$ denote a lithology classification dataset, where $X^{(i)}\in\mathbb{R}^{T_i\times d}$ is a multivariate log sequence with $T_i$ depth samples and $d$ logging channels, and $y^{(i)}=[y^{(i)}_{1},\ldots,y^{(i)}_{T_i}]$ is the corresponding lithology label sequence with $y^{(i)}_{t}\in\mathcal{C}$ for a fixed set of classes $\mathcal{C}$.
The goal is to learn a depth-aligned sequence labeling function $f_{\theta}:\mathbb{R}^{T_i\times d}\rightarrow \mathcal{C}^{T_i}$ that maps each well-log sequence to a lithology prediction at every depth sample, i.e., $\hat{y}^{(i)}=f_{\theta}(X^{(i)})$ with $\hat{y}^{(i)}_{t}\in\mathcal{C}$.
In practice, $f_{\theta}$ may be trained and evaluated using local windows $X^{(i)}_{t:t+k-1}\in\mathbb{R}^{k\times d}$ (with $k\ll T_i$) to handle long sequences, while ensuring that the final predictions remain consistent and aligned over the full depth interval.

\vspace{-0.04in}
\subsection{Overview of the GeoMind Framework}
GeoMind is a lithology classification framework that integrates specialized numerical predictors with large language model-based reasoning through a structured, process-supervised workflow. As illustrated in Figure~\ref{fig:main_fig}, GeoMind organizes lithology interpretation as a multi-step decision process coordinated by a Planner-Executor- Reflector architecture. The planner decomposes the classification task and orchestrates tool usage based on logging data characteristics, the executor performs perception, reasoning, and validation through diverse modular analytical tools, and the reflector resolves conflicts and refines predictions under geological constraints. To enhance domain reliability, the underlying language model is first adapted via supervised fine-tuning on lithology well-log–specific instruction data, and the entire workflow is further optimized using multi-objective process supervision that provides fine-grained rewards at key intermediate reasoning stages.

\subsection{Tool Preparation}
To bridge the gap between raw geophysical signals and geological reasoning, we develop a modular toolkit for evidence-based modeling. It comprises three hierarchical layers: perception tools convert numerical sequences into qualitative trends, reasoning engines integrate multi-source evidence to form lithological hypotheses, and analytical validators audit these hypotheses against stratigraphic constraints to ensure geological consistency. This decoupled architecture enables GeoMind to perform iterative, verifiable decision-making that mirrors human expert analysis. Table~\ref{tab:executor_tools} and Appendix~\ref{app:executor_tools} provide further details on the toolset and its implementation.

\vspace{-0.04in}
\subsection{Agentic Workflow Design} \label{sec: Fragvalid}

We formalize the lithology identification process as an agentic workflow consisting of three distinct phases.

\subsubsection{Planner}

The planner leverages a LLM to perform task decomposition and tool orchestration. Given a well-log segment, the planner analyzes data characteristics such as signal variability and depth continuity, and dynamically generates an execution plan specifying which tools to invoke. Formally, given a well-log window $X_{t:t+k-1}$ and optional context, the planner produces an execution plan as an ordered tool-call sequence:
$
\mathcal{P}_t = (a_{t,1}, a_{t,2}, \ldots, a_{t,L_t}),\quad a_{t,\ell}\in\mathcal{A}(\mathcal{T}),
$
where $\mathcal{A}(\mathcal{T})$ is the action space induced by tool invocations (tool choice and arguments).
This design allows GeoMind to adapt its analysis strategy across different wells and stratigraphic settings, rather than relying on a fixed inference pipeline.

\begin{table}[t]\centering
    \caption{Tools used by the executor and their roles in generating intermediate evidence, candidate predictions, and geological consistency signals.}
    \vspace{-0.08in}
	\scalebox{0.77}{
		\renewcommand{\arraystretch}{1}
		\begin{tabular}{>{\centering\arraybackslash}m{0.1\textwidth} >{\raggedright\arraybackslash}m{0.47\textwidth}}
			\toprule
			\textbf{Tool} & \textbf{Descriptions} \\
			\midrule
			Case Retriever &
			Retrieves $K$ most similar historical windows and their labels as neighborhood evidence. It outputs neighbor tuples $\mathcal{N}_t=\{(X^{(j)},y^{(j)},s_{t,j})\}_{j=1}^{K}$, where $s_{t,j}$ is a normalized similarity score computed from multiple distance metrics. \\
			\midrule
			Trend Pattern Extractor &
			Converts multivariate log curves into a structured natural-language trend description $z_t$, summarizing depth-aware cues such as stable segments, turning points, and magnitude changes to bridge numerical patterns and semantic reasoning. \\
			\midrule
			Neighbor Vote Aggregator &
			Transforms retrieved neighbors into class confidences $p^{\mathrm{nbr}}_t$ via similarity-weighted voting under a local smoothness assumption, providing an empirical prior over lithology classes. \\
			\midrule
			Neural Probability Interpreter &
			Interprets the neural classifier output $p^{\mathrm{nn}}_t$ into concise, human-readable statements (optionally thresholded) and produces an explanation $e_t$ to improve transparency and calibration. \\
			\midrule
			Semantic Reasoning Engine &
			Fuses the well-log table, trend narrative $z_t$, and neighbor evidence $(\mathcal{N}_t, p^{\mathrm{nbr}}_t)$ to generate a stepwise reasoning trace $h_t$ and an LLM candidate prediction $\tilde{y}^{\mathrm{llm}}_t$ (or a distribution over classes). \\
			\midrule
			Consensus Conflict Scanner &
			Compares three predictors (neighbor vote, neural net, and semantic reasoning) to quantify agreement and produce a conflict report $q_t$, highlighting uncertainty regions and disagreement patterns that often occur near boundaries or transitions. \\
			\midrule
			Stratigraphic Sequence Validator &
			Validates plausibility of predicted lithology sequences using a Markov transition model learned from training labels, and outputs a validation signal $v$ that flags low-probability jumps. \\
			\bottomrule
		\end{tabular}
	}
    \vspace{-0.2in}
    \label{tab:executor_tools}
    
\end{table}

\subsubsection{Executor}

Guided by $\mathcal{P}_t$, the executor executes a tool-augmented analysis to produce intermediate evidence and candidate lithology predictions for the current well-log window. Conceptually, this stage forms a three-layer pipeline: \emph{perception} tools transform raw multivariate logs into retrieval- and text-based representations, \emph{reasoning} tools fuse multi-source evidence to yield candidate labels and calibrated confidence signals, and \emph{analysis} tools diagnose inconsistencies and enforce geological plausibility through cross-predictor conflict checks and stratigraphic transition validation.
% Table~\ref{tab:executor_tools} summarizes the tool set, their inputs/outputs, and their roles in the workflow. The concrete formulations and implementation details of each tool are provided in Appendix~\ref{app:executor_tools}.

\subsubsection{Reflector}
% The reflector serves as a critical meta-decision module that systematically consolidates diverse executor outputs. It jointly considers candidate predictions, confidence distributions, conflict analysis reports, and stratigraphic validation results to resolve internal disagreements and produce the final lithology label sequence.
% Formally, we write this aggregation abstractly as:
% $
% (\hat{y}_t, \;\hat{h}_t) = \mathrm{Reflect}\Big(\tilde{y}^{\mathrm{nbr}}_t,\tilde{y}^{\mathrm{nn}}_t,\tilde{y}^{\mathrm{llm}}_t, p^{\mathrm{nbr}}_t, p^{\mathrm{nn}}_t, h_t, q_t, v \Big),
% $
% where $\hat{h}_t$ is the final explanation that integrates evidence and stratigraphic constraints.
% Through reflection, the reflector can revise intermediate conclusions, ensuring that the final prediction is both statistically supported and geologically coherent, while maintaining a transparent and traceable reasoning chain.

The Reflector serves as a critical meta-decision module that systematically consolidates diverse executor outputs. Formally, we define this aggregation process as:
\begin{equation}
\setlength{\abovedisplayskip}{3pt}
\setlength{\belowdisplayskip}{3pt}
    (\hat{y}_t, \hat{h}_t) = \mathrm{Reflect}\Big(
    \underbrace{\tilde{y}^{\mathrm{nbr}}_t, \tilde{y}^{\mathrm{nn}}_t, \tilde{y}^{\mathrm{llm}}_t}_{\text{Candidates}}, \;
    \underbrace{p^{\mathrm{nbr}}_t, p^{\mathrm{nn}}_t}_{\text{Confidences}}, \;
    \underbrace{h_t, q_t, v_t}_{\text{Diagnostics}}
    \Big),
\end{equation}
where the inputs consist of: 
(i) \textbf{Candidate predictions} from neighbor voting ($\tilde{y}^{\mathrm{nbr}}_t$), the neural classifier ($\tilde{y}^{\mathrm{nn}}_t$), and LLM reasoning ($\tilde{y}^{\mathrm{llm}}_t$); 
(ii) \textbf{Confidence distributions} ($p^{\mathrm{nbr}}_t, p^{\mathrm{nn}}_t$) indicating statistical certainty; and 
(iii) \textbf{Diagnostic context}, including the reasoning trace $h_t$, the conflict report $q_t$, and the stratigraphic validation signal $v_t$. 
The Reflector synthesizes these signals to produce the final lithology label $\hat{y}_t$ and a refined explanation $\hat{h}_t$, ensuring the prediction is both statistically supported and geologically coherent.

\vspace{-0.1in}
\subsection{Training Strategy}

\subsubsection{Domain Adaptive Supervised Fine-Tuning} \label{sec:sft}

To equip the LLM with lithology-specific priors and well-log interpretation conventions, we perform domain-adaptive supervised fine-tuning on Qwen3-4B. We curate an instruction set by consolidating lithology knowledge from resources such as Wikipedia-style descriptions and restructuring it into actionable QA-style supervision; to improve data quality and reduce single-teacher bias, the SFT instructions are constructed offline by teacher models (Gemini-3-Pro) using the training split. Specifically, we distill common lithology identification cues, log-curve interpretations, canonical diagnostic criteria, and handling heuristics into 600 QA, and further add 200 cross-type contrastive QA samples (e.g., sandstone vs.\ siltstone, shale vs.\ mudstone, carbonate vs.\ clastic) to inject analysis habits that align the LLM with supervised lithology labeling.
This contrastive supervision enforces a comparative reasoning paradigm, guiding the model to move beyond brittle range-rule judgments. Instead, it learns to summarize global characteristics and distinct boundary cues, which effectively mitigates the impact of logging noise and aligns the model's predictions with invariant geological semantics.
% The contrastive instructions constrain the model to move beyond brittle range-rule judgments and instead summarize global characteristics, trend patterns, and discriminative differences, encouraging reasoning that better matches how labels are defined under distribution shift and logging heterogeneity, thereby improving label semantic alignment and robustness to inter-well variations and log measurements.

\begin{figure*}[t] 
    \centering
    \includegraphics[width=0.96\textwidth]{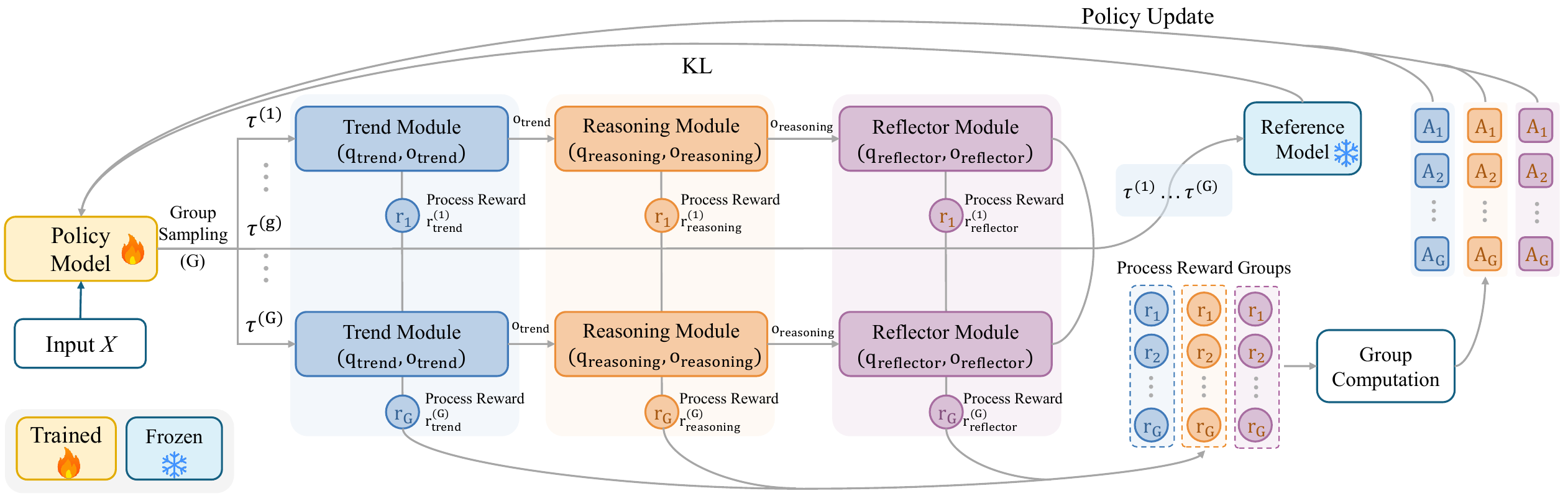} 
    \vspace{-0.15in}
    \caption{MAPO architecture with group sampling, module-specific process rewards, and KL-regularized policy update.} 
    \label{fig:main_fig_magrpo} 

\end{figure*}
\subsubsection{Process Reward Design}\label{sec:process_reward}

Unlike conventional approaches that supervise only final classification outcomes, GeoMind injects \emph{process rewards} at key intermediate stages to guide the entire reasoning workflow.
We design three complementary reward components.

\paragraph{Trend Analysis Reward}
To provide privacy-preserving, stable, and rubric-consistent process rewards, we distill multi-criteria judgments into a local trend reward model.
We use GPT-5 only as an \emph{offline teacher} to construct supervised data and fine-tune a Qwen3-4B evaluator.
From training wells, we sample fixed-length windows $X_t$.
For each $X_t$, we query GPT-5 under two prompting paradigms: (a) \emph{describe-then-classify}, which first produces a trend narrative $z_t$ and then predicts the label, and (b) \emph{direct classification}.
We repeat inference multiple times with stochastic decoding and compute $\mathrm{pass@}k$ accuracy for both paradigms.
A window is regarded as \emph{trend-helpful} if the narrative improves decision quality:
\begin{equation}
\Delta_t = \mathrm{pass@}k_{\mathrm{trend}}(X_t) - \mathrm{pass@}k_{\mathrm{direct}}(X_t) \;>\; 0.
\label{eq:trend_helpful}
\end{equation}
We focus reward-model supervision on these windows during training so that the learned reward emphasizes decision-useful narratives rather than stylistic preferences.

For each selected window $X_t$, we treat the teacher-generated narrative $z_t$ as a high-quality reference and create diagnostically incorrect narratives via structured perturbations.
We generate perturbed curves $X_t'$ by applying controlled operations (e.g., shifting turning points, flipping trend directions, or altering depth alignment), and ask GPT-5 to narrate $X_t'$ to obtain $z_t'$.
We then form \emph{mismatched} training pairs $(X_t, z_t')$ so that $z_t'$ serves as a realistic hard negative for the original evidence $X_t$.
Finally, GPT-5 assigns rubric-based scores on four criteria---Accuracy, Completeness, Clarity, and Depth alignment.
We employ standard supervised fine-tuning on Qwen3-4B to output the four rubric scores. The final trend reward is computed as the equally weighted sum of the four normalized rubric scores and the terminal lithology accuracy.
% At runtime, the trend reward is computed locally as:
% \begin{equation}
% \big(\hat{r}^{\mathrm{acc}}_t,\hat{r}^{\mathrm{com}}_t,\hat{r}^{\mathrm{cla}}_t,\hat{r}^{\mathrm{dep}}_t\big)
% = f_{\phi}(X_t, z_t),
% \label{eq:trend_rm}
% \end{equation}
% \begin{equation}
% R^{\mathrm{trend}}_t
% = \hat{r}^{\mathrm{acc}}_t + \hat{r}^{\mathrm{com}}_t + \hat{r}^{\mathrm{cla}}_t + \hat{r}^{\mathrm{dep}}_t.
% \label{eq:trend_reward}
% \end{equation}
This process reward encourages faithful and complete abstractions of well logs and keeps RL training fully self-contained.

\paragraph{LLM Classification Accuracy Reward}
After the semantic reasoning engine outputs $\tilde{y}^{\mathrm{llm}}$ for a window, we measure prediction correctness against ground truth.
For a window of length $k$ (or any evaluation batch), the accuracy reward is:
\begin{equation}
R^{\mathrm{llm\_acc}} = \frac{1}{k}\sum_{u=1}^{k} \mathbb{I}\!\left[\tilde{y}^{\mathrm{llm}}_{u} = y_{u}\right] \in [0,1].
\label{eq:llm_acc_reward}
\end{equation}
\noindent This provides supervision for intermediate reasoning outputs, helping stabilize training and improve workflow convergence.

\paragraph{Reflection Correction Reward}
The reflector is rewarded for \emph{resolving predictor conflicts by selecting the correct label when evidence permits}.
Let the candidate labels be
$\tilde{y}^{(1)}_t=\tilde{y}^{\mathrm{nbr}}_t$, $\tilde{y}^{(2)}_t=\tilde{y}^{\mathrm{nn}}_t$, and $\tilde{y}^{(3)}_t=\tilde{y}^{\mathrm{llm}}_t$,
and denote the candidate set as $\mathcal{Y}_t=\{\tilde{y}^{(1)}_t,\tilde{y}^{(2)}_t,\tilde{y}^{(3)}_t\}$.
We quantify the level of agreement among predictors by
\begin{equation}
m_t \;=\; \max_{c\in\mathcal{C}} \sum_{i=1}^{3}\mathbb{I}\!\left[\tilde{y}^{(i)}_t=c\right],
\label{eq:refl_agreement_level}
\end{equation}
\noindent where $m_t\in\{1,2,3\}$ indicates whether the candidates are fully split ($m_t=1$), partially aligned ($m_t=2$), or unanimous ($m_t=3$).

% Our reward emphasizes \emph{actionable correction}: when the ground-truth label is present among candidates (i.e., $y_t\in\mathcal{Y}_t$), the reflector should identify and select it; otherwise, the step provides no direct correction signal, and we avoid injecting noisy supervision.

The Reflector predicts over $\mathcal{C}$, whereas $R_t^{\mathrm{refl}}$ supervises only candidate-resolvable cases ($y_t\in\mathcal{Y}_t$); other cases rely on terminal correctness.
Meanwhile, when all predictors are unanimous ($m_t=3$), choosing that consensus is treated as a baseline behavior and does not receive additional reward.
This yields the compact form
\begin{equation}
R^{\mathrm{refl}}_t
=
\mathbb{I}[y_t\in\mathcal{Y}_t]\,
\mathbb{I}[m_t<3]\cdot
\left(
\mathbb{I}[\hat{y}_t=y_t]
-
\mathbb{I}[\hat{y}_t\neq y_t]\eta(m_t)
\right).
\label{eq:refl_compact}
\end{equation}
where $\hat{y}_t$ is the reflector's decision and $\eta(m_t)$ modulates the penalty by disagreement severity.
In particular, we set $\eta(2)=1.0$ to penalize missing an available correction when two predictors align, and $\eta(1)=0.5$ to apply a milder penalty under maximal conflict, reflecting the higher intrinsic difficulty.

\subsubsection{MAPO}
Training agentic frameworks with multi-step reasoning capabilities presents a significant challenge: standard RL approaches typically assign a single scalar reward at the trajectory's termination (e.g., accuracy). In a complex agentic workflow, this sparse signal exacerbates the \textit{credit assignment problem}, where the policy struggles to distinguish which specific module contributed to the success or failure, such sparse rewards lead to poor convergence and difficulty in credit assignment, as a correct final prediction might be reached through flawed intermediate reasoning.
To address this, we propose \textbf{module-aware group relative policy optimization (MAPO)}. In Figure~\ref{fig:main_fig_magrpo}, MAPO treats the agentic workflow as a composite of distinct, independently optimizable interaction events. By leveraging the process rewards defined in Sec.~\ref{sec:process_reward}, we compute advantages for each module locally within a group of sampled trajectories, ensuring that each component of the agent is supervised by its most relevant objective.

\paragraph{Formulation}
Let the GeoMind workflow be represented as a sequential generation process involving a set of trainable modules $\mathcal{M}=\{\text{Planner},\text{Trend},\text{Reasoning},\text{Reflector}\}$. 
The Planner is optimized using the terminal correctness reward $R_{\mathrm{final}}$. For a given input well-log window $X$, the agent generates a global trajectory $\tau$ consisting of ordered query-response pairs $\tau = \{(q_m, o_m)\}_{m \in \mathcal{M}}$, where $q_m$ is the prompt context for module $m$ (which may depend on previous outputs) and $o_m$ is the generated response.

We employ a group sampling mechanism to estimate baselines. For each training step, given an input $X$, we rollout a group of $G$ independent global trajectories $\{\tau^{(g)}\}_{g=1}^G$.
The trajectories diverge after the first step. The $g$-th trajectory is a consistent causal chain:
$
\tau^{(g)} = \left( (q_{\text{trend}}^{(g)}, o_{\text{trend}}^{(g)}), \dots, (q_{\text{refl}}^{(g)}, o_{\text{refl}}^{(g)}) \right).
$
This sampling pattern ensures that we capture the variance in reasoning paths while maintaining continuity and consistency within each group member.

\paragraph{Module-Specific Advantage Estimation.}
Instead of aggregating rewards into a single trajectory return, we assign the specific process rewards derived in Sec.~\ref{sec:process_reward} to their corresponding module interactions. For a module $m$ in trajectory $g$, let $r_{m}^{(g)}$ denote the obtained process reward (e.g., $R^{\mathrm{trend}}$ for the Trend Narrator). We compute the advantage $A_{m}^{(g)}$ by normalizing the reward against the group statistics for that specific module step:
\begin{equation}
\setlength{\abovedisplayskip}{3pt}
\setlength{\belowdisplayskip}{3pt}
A_{m}^{(g)} = \frac{r_{m}^{(g)} - \mu_{m}}{\sigma_{m} + \epsilon},
\label{eq:group_advantage}
\end{equation}
where $\mu_{m} = \frac{1}{G}\sum_{i=1}^G r_{m}^{(i)}$ and $\sigma_{m} = \sqrt{\frac{1}{G}\sum_{i=1}^G (r_{m}^{(i)} - \mu_{m})^2}$ are the mean and standard deviation of the rewards for module $m$.

To account for state heterogeneity, we represent each module prompt as $h_m^{(g)}=\psi(q_m^{(g)})$ and measure its dispersion as $d_m=G^{-1}\sum_g|h_m^{(g)}-\bar h_m|_2^2$, where $\bar h_m=G^{-1}\sum_g h_m^{(g)}$. We apply a state gate $\alpha_m=(1+\gamma d_m)^{-1}$ and use $\widetilde{A}_m^{(g)}=\alpha_m A_m^{(g)}$. The gate preserves updates when module actions are generated from similar contexts and attenuates updates when their contexts have diverged.

\vspace{-0.06in}
\paragraph{Joint Optimization Objective.}

A key distinction from standard GRPO is that MAPO \emph{does not} treat the workflow as a single trajectory with a shared return. Instead, each LLM call $(q_m,o_m)$ is optimized with a \emph{module-local} objective using its attenuated group-relative advantage $\widetilde{A}_m^{(g)}$. Concretely, we form a module-indexed training set of interaction events by collecting the $m$-th query--response pairs across the group, and apply a GRPO-style update \emph{only} to the tokens generated within that module. This provides module-aware credit assignment: process rewards supervise the module that generated the corresponding intermediate artifact, while state-dispersed groups receive smaller updates.

Formally, for each module $m\in\mathcal{M}$ and each sampled global rollout $g$, define the module interaction event $e_m^{(g)}=(q_m^{(g)},o_m^{(g)},r_m^{(g)})$. MAPO maximizes the \emph{module-conditioned} surrogate objectives:
\begin{equation}
\setlength{\abovedisplayskip}{3pt}
\setlength{\belowdisplayskip}{3pt}
\begin{aligned}
\mathcal{J}_{\text{MAPO}}(\theta)
&=\sum_{m\in\mathcal{M}}
\mathbb{E}_{q_m \sim \mathcal{D}_m}\Bigg[
\frac{1}{G}\sum_{g=1}^{G}
\Big(
\rho_{m}^{(g)}(\theta)\,\widetilde{A}_{m}^{(g)}
\\
& \hspace{-5mm} \;-\;\beta\,\mathbb{D}_{\mathrm{KL}}\!\left(\pi_{\theta}(\cdot\!\mid q_m)\,\|\,\pi_{\text{ref}}(\cdot\!\mid q_m)\right)
\Big)
\Bigg],
\label{eq:ma_grpo_obj_module}
\end{aligned}
\end{equation}
where $\rho_{m}^{(g)}(\theta)=\frac{\pi_{\theta}(o_m^{(g)}\mid q_m^{(g)})}{\pi_{\theta_{\text{old}}}(o_m^{(g)}\mid q_m^{(g)})}$ is the probability ratio for \emph{module} $m$, and $\mathcal{D}_m$ denotes the distribution of module prompts induced by running the workflow (i.e., the prompt pool for module $m$). The state-dispersion factor preserves module-wise process-reward routing while reducing updates driven by cross-state comparisons.

\vspace{-0.06in}
\section{Experiments}

\subsection{Experimental Setup}

\subsubsection{Datasets.}
We benchmark our approach on four public well-log datasets covering diverse geological settings to evaluate sequence labeling performance: \textbf{SEAM}\footnote{\url{https://wiki.seg.org/wiki/Open_data\#Well_logs}}, from the SEG Wiki portal, serves as a standard benchmark providing depth-aligned multichannel logs with lithology labels.
\textbf{Facies}\footnote{\url{https://www.kaggle.com/datasets/imeintanis/well-log-facies-dataset}} contains logs from the Council Grove gas reservoir (Kansas), with labels derived from core interpretations.
\textbf{FORCE}\footnote{\url{https://www.kaggle.com/datasets/faresazzam/well-logs-dataset-for-machine-learning}} is designed for lithology prediction using multiple geophysical measurements.
\textbf{GeoLink}\footnote{\url{https://github.com/LukasMosser/geolink_dataset?tab=readme-ov-file}}: a large-scale North Sea collection by GEOLINK-S2, featuring depth-aligned lithology interpretations and wireline logs.

% Preamble required: \usepackage{booktabs}
\begin{table}[t]
\centering
\large
\caption{Statistics of the datasets used in this work. Well counts are reported as train/validation/test; 'Interval' denotes the sampling interval.}
\vspace{-0.08in}
\label{tab:dataset_stats}
\setlength{\tabcolsep}{8pt} % 稍微调整列间距，使表格更舒展
\renewcommand{\arraystretch}{1.15}
\resizebox{0.48\textwidth}{!}{
\begin{tabular}{cccccc} % 所有数字列右对齐(r)，视觉上更整齐
\toprule
\textbf{Dataset} & \textbf{\# Wells (Tr./Va./Te.)} & \textbf{\# Samples} & \textbf{\# Channels} & \textbf{\# Classes} & \textbf{Interval} \\
\midrule
SEAM    & 3/1/1   & 7,092   & 9  & 7  & 10m    \\
Facies  & 4/1/2   & 3,164   & 7  & 9  & 0.5m   \\
FORCE   & 4/2/5   & 52,766  & 9  & 5  & 0.15m  \\
Geolink & 80/23/25 & 580,205 & 9  & 11 & 0.125m \\
\bottomrule
\end{tabular}
}
\vspace{-0.22in}
\end{table}

\begin{table*}[t]
\centering
\caption{Performance comparison on four lithology-related datasets. Best results are in \textbf{bold} and second-best are \underline{underlined}.}
\vspace{-0.08in}
% --- 颜色与间距设置 ---
\definecolor{highlight}{HTML}{EFEFEF}
\setlength{\aboverulesep}{0pt} % 消除白缝，确保rowcolor填满
\setlength{\belowrulesep}{0pt}
\setlength{\extrarowheight}{.95ex} % 增加行高以补偿视觉
\setlength{\tabcolsep}{1pt}      % 紧凑列宽
\renewcommand{\arraystretch}{0.9}
% --- 定义等宽居中列 ---
\newcolumntype{Y}{>{\centering\arraybackslash}X}

\resizebox{\textwidth}{!}{%
\begin{tabularx}{\textwidth}{l *{12}{Y}} 
\toprule
\multirow{2}{*}{Method} & 
\multicolumn{3}{c}{SEAM} & \multicolumn{3}{c}{Facies} & \multicolumn{3}{c}{FORCE} & \multicolumn{3}{c}{GeoLink} \\
\cmidrule(lr){2-4} \cmidrule(lr){5-7} \cmidrule(lr){8-10} \cmidrule(lr){11-13}
& Precision & Recall & F1 & Precision & Recall & F1 & Precision & Recall & F1 & Precision & Recall & F1 \\
\midrule

nn-DTW         & 0.8112 & 0.8276 & 0.8243 & 0.3705 & 0.3565 & 0.3487 & 0.3213 & 0.3200 & 0.2905 & 0.3342 & 0.3847 & 0.3479 \\
GBDT          & 0.8184 & 0.8522 & 0.8393 & 0.4673 & 0.3951 & 0.3702 & 0.2293 & 0.2712 & 0.2501 & 0.4251 & 0.4029 & 0.4122 \\
XGBoost       & 0.8379 & \textbf{0.8707} & 0.8355 & \textbf{0.4725} & 0.4004 & 0.4086 & \underline{0.4329} & 0.3718 & 0.3612 & 0.3825 & 0.3623 & 0.3644 \\
LSTMFCN       & 0.8120 & 0.8590 & 0.8312 & 0.3261 & 0.3512 & 0.3270 & 0.3030 & 0.3441 & 0.2907 & 0.3729 & 0.3764 & 0.3705 \\
MLP           & 0.8364 & 0.8592 & 0.8426 & 0.2841 & 0.3394 & 0.3067 & 0.2647 & 0.3074 & 0.2769 & 0.3894 & 0.4141 & 0.4054 \\
InceptionTime & \underline{0.8535} & 0.8578 & 0.8537 & 0.4505 & 0.4069 & 0.4062 & 0.3931 & \textbf{0.4054} & \underline{0.3614} & 0.4170 & 0.4125 & 0.4116 \\
MiniRocket    & 0.8442 & 0.8607 & \underline{0.8583} & 0.3676 & 0.3223 & 0.2859 & 0.3221 & 0.3014 & 0.2876 & 0.3948 & 0.4095 & 0.4027 \\
MOMENT        & 0.8492 & 0.8609 & 0.8515 & 0.3582 & 0.3964 & 0.3714 & 0.3478 & 0.3448 & 0.3438 & 0.3715 & 0.4078 & 0.3850 \\
BiGRU-CRF     & 0.8337 & 0.8492 & 0.8410 & 0.3978 & 0.4251 & 0.4110 & 0.3394 & 0.3531 & 0.3460 & 0.3945 & 0.4097 & 0.4020 \\
Inception-Geol & 0.8416 & 0.8565 & 0.8490 & 0.4108 & 0.4338 & 0.4220 & 0.3442 & 0.3577 & 0.3510 & 0.4034 & 0.4187 & 0.4110 \\
GIAT          & 0.8461 & 0.8597 & 0.8530 & 0.4136 & 0.4367 & \underline{0.4250} & 0.3398 & 0.3527 & 0.3460 & 0.4118 & 0.4285 & \underline{0.4200} \\
InstructTime  & 0.8389 & 0.8553 & 0.8428 & 0.3854 & 0.4017 & 0.3896 & 0.2815 & 0.3279 & 0.2994 & 0.3647 & 0.3972 & 0.3768 \\
GPT4TS        & 0.8486 & 0.8467 & 0.8425 & 0.4439 & \underline{0.4325} & 0.4155 & 0.3851 & 0.3784 & 0.3453 & \underline{0.4291} & \underline{0.4161} & 0.4130 \\
TableTime     & 0.8335 & 0.8419 & 0.8367 & 0.3529 & 0.3729 & 0.3615 & 0.2794 & 0.3347 & 0.3017 & 0.3382 & 0.3916 & 0.3591 \\

\midrule

\rowcolor{highlight} 
\textbf{Ours} & \textbf{0.8827} & \underline{0.8814} & \textbf{0.8821} & \underline{0.4837} & \textbf{0.4708} & \textbf{0.4582} & \textbf{0.4436} & \underline{0.4018} & \textbf{0.3741} & \textbf{0.4521} & \textbf{0.4393} & \textbf{0.4401} \\
\bottomrule
\end{tabularx}%
}

\vspace{-0.1in}
\label{tab:main_results}
\end{table*}
\subsubsection{Baselines.}
To conduct a comprehensive comparison, we compare against several baselines: \textbf{Machine learning-based} methods: nn-DTW~\cite{kate2016using}, GBDT~\cite{ke2017lightgbm}, XGBoost~\cite{chen2015xgboost}.
\textbf{Deep learning-based} methods: LSTMFCN~\cite{karim2019multivariate}, MLP~\cite{taud2017multilayer}, InceptionTime~\cite{ismail2020inceptiontime}, MiniRocket~\cite{dempster2021minirocket}, MOMENT~\cite{goswami2024moment}.
\textbf{Domain-specific lithology} methods: BiGRU-CRF~\cite{liu2021lithological}, Inception-Geol~\cite{tschannen2017facies}, and GIAT~\cite{li2026giat}.
\textbf{LLM-based} approaches: InstructTime~\cite{cheng2025instructime}, GPT4TS~\cite{zhou2023one} and TableTime~\cite{wang2025tabletime}.

\subsubsection{Evaluation Metrics.} 
To evaluate the lithology classification performances, here we select three widely used metrics~\cite{zheng2014time,tonutti2019robust}, i.e., Precision, Recall ~\cite{liu2011enhancing}, and F1 measure~\cite{li2021cross}. Considering our task is a multi-class classification problem, we use the weighted average scores to evaluate the performances of our proposed methods and all baselines. Specifically, we weight the metrics of each class by the number of samples from that class.

\subsubsection{Implementation Details}
Our GeoMind framework was trained and evaluated on a machine equipped with 4$\times$ NVIDIA A800 GPUs, due to the additional memory and compute overhead introduced by multi-stage tool invocation and long-horizon rollouts. For the numerical module in GeoMind, we use GPT4TS as the numerical predictor, with the same input preprocessing protocol as other neural baselines. We report the mean performance over multiple runs with different random seeds. More detailed deployment specifications can be found in Appendix \ref{imp_appendix}.

\begin{table}[t]
\centering
% \small
% 调整行距和列间距，适应双栏排版
\renewcommand{\arraystretch}{1}
\setlength{\tabcolsep}{3pt}
\caption{Ablation study of the GeoMind framework. We report the F1 score across four benchmarks. The notation ``maj. vote'' refers to the majority voting strategy.}
\vspace{-0.08in}
\begin{tabular}{
    @{} l % 左对齐，去掉左侧空白
    S[table-format=1.4]
    S[table-format=1.4]
    S[table-format=1.4]
    S[table-format=1.4]
    @{} % 去掉右侧空白
}
\toprule
\textbf{Method} & {\textbf{SEAM}} & {\textbf{Facies}} & {\textbf{FORCE}} & {\textbf{GeoLink}} \\
\midrule
% 主模型高亮
\textbf{GeoMind(Ours)} & \textbf{0.8821} & \textbf{0.4582} & \textbf{0.3741} & \textbf{0.4401} \\
\midrule
% 第一组消融：训练策略
\multicolumn{5}{l}{\textit{Impact of Training Strategy}} \\
\hspace{1em} w/o Domain-adaptive SFT  & 0.8756 & 0.4521 & 0.3693 & 0.4348 \\
\hspace{1em} w/o Agentic Workflow RL  & 0.8157 & 0.3869 & 0.3128 & 0.3617 \\
\hspace{1em} w/o Process Rewards      & 0.8684 & 0.4417 & 0.3573 & 0.4256 \\
\hspace{1em} w/o MAPO (std. GRPO)             & 0.8619 & 0.4347 & 0.3518 & 0.4189 \\
\midrule
% 第二组消融：Agent 组件
\multicolumn{5}{l}{\textit{Impact of Agentic Components}} \\
\hspace{1em} w/o Reflector (maj. vote) & 0.8708 & 0.4463 & 0.3624 & 0.4283 \\
\hspace{1em} w/o Reflector (LLM only)  & 0.8667 & 0.4389 & 0.3541 & 0.4247 \\
\hspace{1em} w/o Planner (fixed plan)  & 0.8742 & 0.4495 & 0.3658 & 0.4316 \\
\bottomrule
\end{tabular}

\vspace{-0.1in}
\label{tab:ablation}
\end{table}
\vspace{-0.04in}
\subsection{Main Results}
% Table~2 summarizes the overall performance on four lithology benchmarks. Across all datasets, GeoMind consistently achieves the strongest (or near-strongest) weighted F1, indicating that the proposed agentic workflow improves not only pointwise classification accuracy but also the global balance between Precision and Recall under class imbalance. Notably, the gains are not confined to a single family of baselines: GeoMind surpasses classical machine-learning models, competitive deep time-series classifiers, and recent LLM-oriented formulations. This suggests that the improvement does not come from simply scaling a backbone model, but from the \emph{coordination mechanism} that reconciles heterogeneous evidence sources (numerical patterns, neighborhood consistency, and semantic/geological reasoning) into a verifiable decision process.

% Beyond aggregate scores, a key observation is that GeoMind's advantage is more pronounced on datasets with more severe distribution shift, inter-well heterogeneity, or boundary ambiguity. In such cases, purely numerical models may overfit local signal statistics, while LLM-only reasoning can be unstable when faced with noisy multi-channel measurements. GeoMind addresses this tension by explicitly exposing intermediate evidence and training the workflow with process supervision. As a result, GeoMind tends to produce more stable predictions around lithological transitions, which is critical for downstream stratigraphic interpretation.

Table \ref{tab:main_results} presents the overall performance comparison across four lithology benchmarks. The results demonstrate that GeoMind consistently achieves superior performance compared to state-of-the-art methods. The key observations are summarized as follows:

\noindent (1) \textbf{Consistent Superiority via Multi-Source Coordination}: Across all datasets, GeoMind consistently achieves the strongest (or near-strongest) weighted F1 scores, effectively balancing the trade-off between precision and recall. Notably, these gains are not limited to a single baseline category; GeoMind surpasses classical machine learning models, deep time-series classifiers, and recent LLM-based formulations. This indicates that the improvement stems not merely from scaling the backbone, but from the proposed coordination mechanism. By reconciling heterogeneous evidence sources: numerical patterns, neighborhood consistency, and semantic reasoning, GeoMind synthesizes a more verifiable and accurate decision process than individual paradigms.

\noindent (2) \textbf{Stability in Complex Geological Settings}: GeoMind’s performance advantage is particularly pronounced on datasets characterized by severe distribution shifts, inter-well heterogeneity, or ambiguous boundaries. In these complex scenarios, purely numerical models often overfit local signal statistics, while standalone LLMs can become unstable due to noisy multi-channel measurements. GeoMind addresses this tension through process supervision, which explicitly exposes intermediate evidence and guides the workflow. This results in significantly more stable predictions around critical lithological transitions, which is essential for reliable downstream stratigraphic interpretation.

\begin{figure*}[t] \centering
    \makebox[0.33\textwidth]{(a) Actor Gradient Norm}
    \makebox[0.33\textwidth]{(b) Training Trajectory Return}
    \makebox[0.33\textwidth]{(c) Validation Trajectory Return}
    \\
    \includegraphics[width=0.33\textwidth]{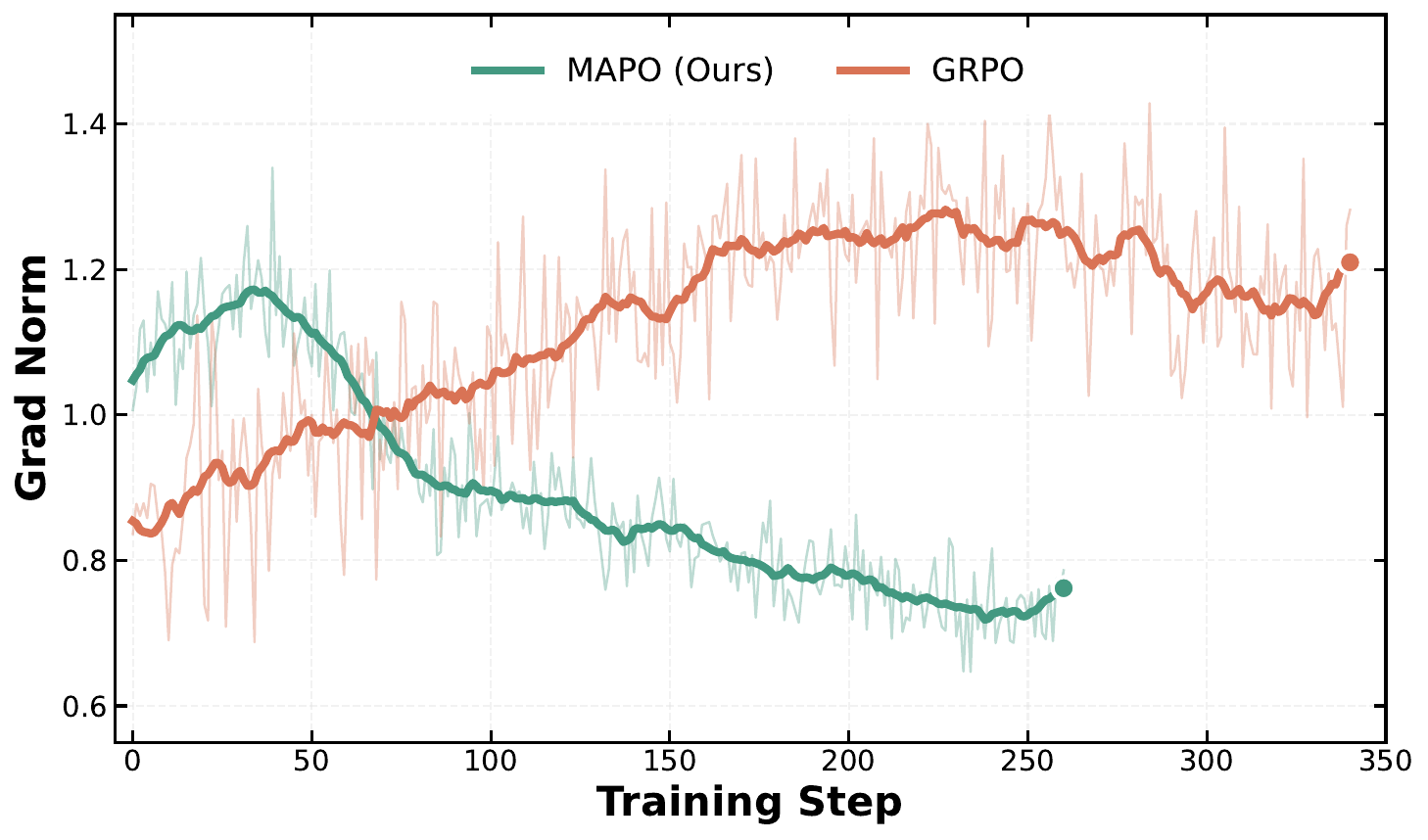}
    \includegraphics[width=0.33\textwidth]{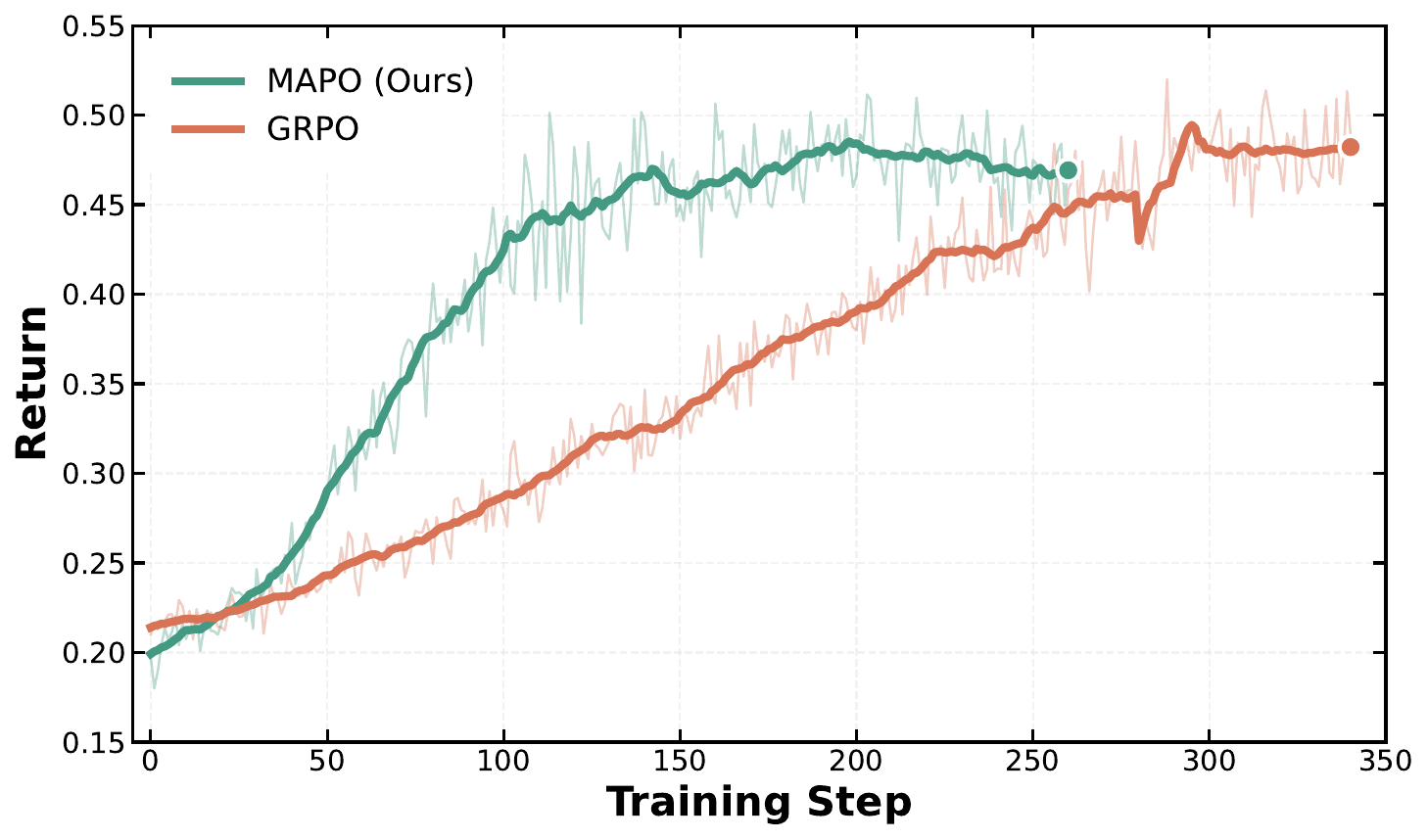}
    \includegraphics[width=0.33\textwidth]{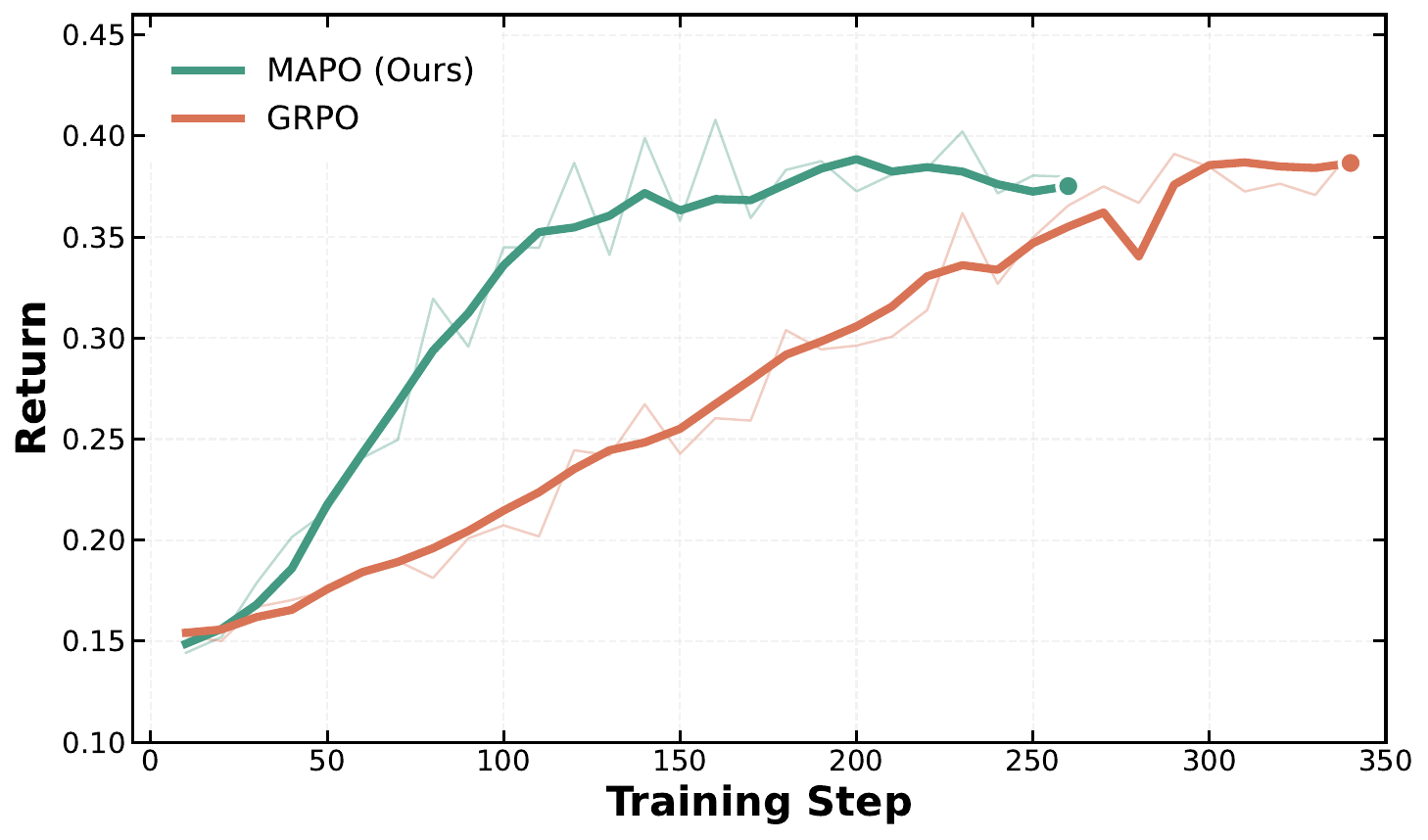}
    \vspace{-0.28in}
    \caption{Comparison of actor gradient norm and trajectory return between MAPO and GRPO. MAPO demonstrates smoother gradient dynamics and faster improvement in trajectory returns, indicating a more stable process-supervised optimization.}
    % \caption{Actor gradient norm and trajectory return for MAPO and GRPO. MAPO shows smoother gradient dynamics and earlier return gains.}
    \vspace{-0.06in}
    \label{fig:comp_grpo}
\end{figure*}

\begin{figure}[t] \centering
    \includegraphics[width=0.48\textwidth]{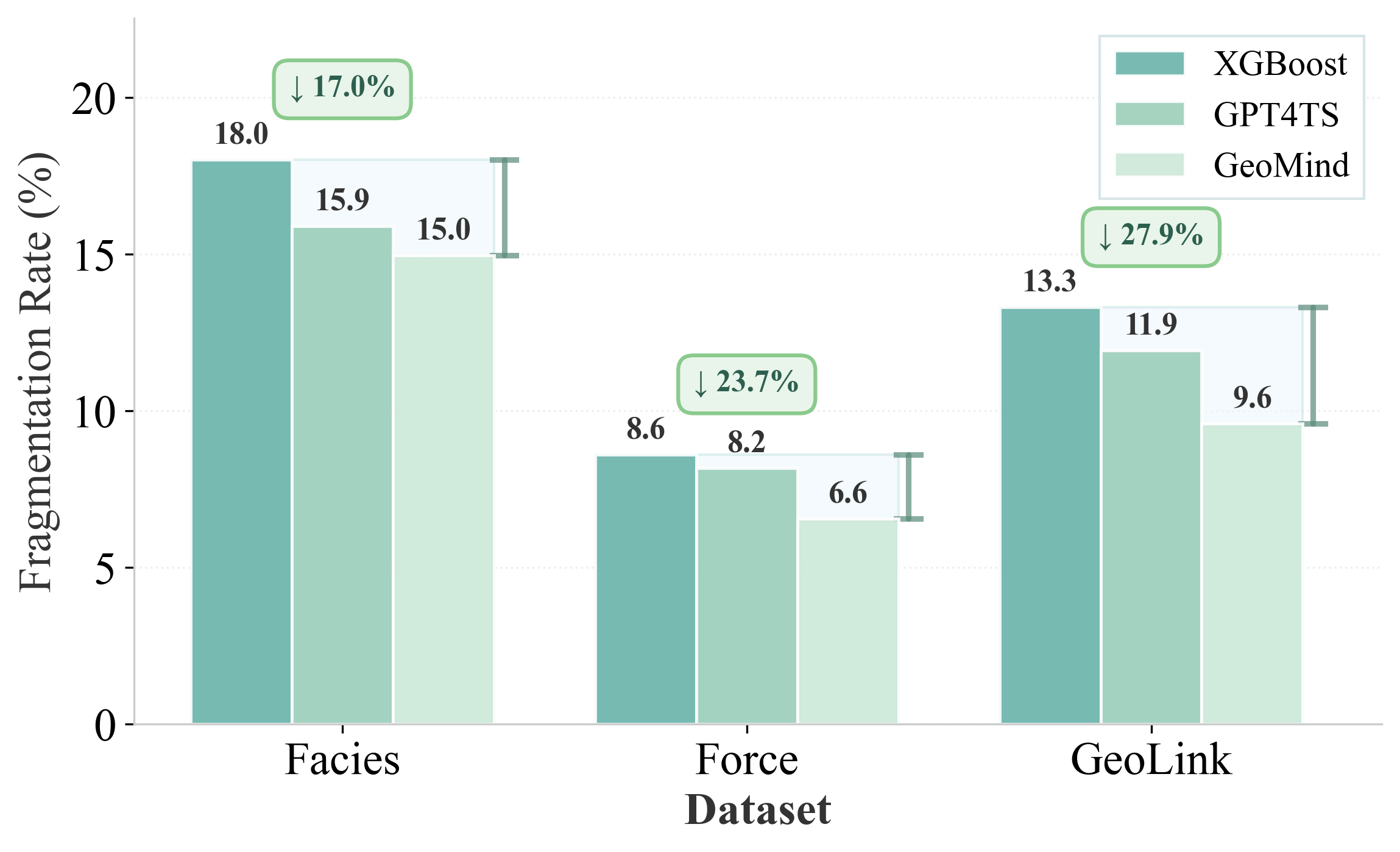}
    \vspace{-0.18in}
    \caption{Comparison of fragmentation rates across Facies, FORCE, and GeoLink datasets. GeoMind consistently achieves the lowest rates, demonstrating a relative reduction of up to 27.9\% compared to the XGBoost baseline.}
    \label{fig:frag}
    \vspace{-0.08in}
\end{figure}

\subsection{Ablation Study}

To validate the contribution of specific components within GeoMind, we perform an ablation study across all four benchmarks. The results, summarized in Table 4, isolate the effects of our training strategies and the modular design.

\paragraph{Impact of Training Strategy.}
The system benefits primarily from learning a workflow-aware policy rather than relying solely on the base model's domain adaptation. Removing the agentic workflow RL (performing inference directly with the SFT model) causes the most significant performance drop. This suggests that end-to-end optimization of multi-step decision-making is critical for interpreting well logs, particularly where boundary ambiguity and distribution shifts occur. Consequently, GeoMind's performance gains stem from coordinating planning, execution, and verification rather than stronger individual step predictions.

Process-level supervision and module-aware optimization also yield measurable improvements. Removing process rewards leads to consistent performance drops, indicating that shaping intermediate reasoning helps stabilize predictions. Replacing MAPO with standard GRPO (using only outcome rewards) similarly degrades results, underscoring the need to mitigate optimization interference and assign credit to specific workflow stages. SFT contributes a smaller but steady gain, providing a domain prior while the primary improvements arise from RL-driven workflow learning.

\paragraph{Impact of Agentic Components.}
We examine the agentic modules to understand the role of orchestration and self-correction. Both the reflector and planner are essential for reliable predictions, with the reflector serving a central role in reconciling conflicting cues and maintaining robustness under uncertainty. Replacing the reflector with majority voting underperforms the full model, as simple aggregation fails to exploit structured feedback. Removing the reflector entirely (the ``LLM-only'' setting) causes further degradation, demonstrating that effective reflection requires integrating multi-source evidence rather than a single reasoning path. Furthermore, using a fixed plan where all tools execute sequentially reduces accuracy. This confirms that adaptive planning, which orders actions based on local signals, is superior to a static pipeline.

\subsection{Analysis of Stratigraphic Fragmentation} \label{sec:fragmentation}

To quantify the physically implausible ``salt-and-pepper'' noise common in discriminative models, we use a class-conditioned short-run fragmentation rate. For each dataset $d$ and lithology class $c$, we estimate a threshold $k_{d,c}$ from training labels only as the 5th percentile of the lengths of maximal contiguous ground-truth runs of class $c$. A maximal predicted run $\hat r$ is counted as a fragment when $|\hat r|<k_{d,c(\hat r)}$, and the rate is the percentage of such runs among all predicted runs. The class-specific lower-tail criterion identifies predictions that are unusually short relative to the observed thickness distribution of the same lithology, without imposing a single absolute cutoff across heterogeneous classes or datasets. As shown in Figure~\ref{fig:frag}, GeoMind consistently produces fewer anomalously short runs than XGBoost and GPT4TS, reducing fragmentation by up to 27.9\% relative to XGBoost. A complementary analysis is provided in Appendix~\ref{app:fragmentation_metric}.

\vspace{-0.06in}
% (3) 换小模型基座：不同 backbone 都稳定提升 柱状图
\subsection{Compatibility Across Base Predictors}
To verify that GeoMind is not coupled to a specific lightweight predictor, we replace the numerical backbone with three representative time-series classifiers (XGBoost, InceptionTime and GPT4TS) and report the F1 on four benchmarks. Table~\ref{tab:backbone_generalization} shows GeoMind yields consistent gains across every backbone. The most pronounced gains occur on the challenging datasets (e.g., FORCE and Facies), where the preliminary results indicate improvements of approximately 5.7 F1 points for XGBoost on FORCE and 3.4 F1 points for InceptionTime on Facies. This pattern suggests that the workflow coordination and process supervision are particularly effective under distribution shifts and boundary ambiguity. Across backbones, GeoMind generally increases F1, suggesting that the workflow stabilizes decision-making by reconciling heterogeneous evidence.

% (6) 过程奖励：收敛更快、更稳、效果更好 三个收敛曲线图
\subsection{Process Rewards for Stable Convergence}
% We examine whether fine-grained process rewards can stabilize RL training for our multi-step Planner--Executor--Reflector workflow by comparing MAPO with standard GRPO.  Figure~3 shows that MAPO drives a faster and smoother decrease in critic loss, and reaches higher trajectory returns earlier on both the training and validation splits, while GRPO improves more slowly with larger fluctuations. These observations are consistent with our design choice to supervise intermediate trend narration, LLM decision accuracy, and reflection-based correction, and to optimize them with module-aware, group-relative advantages to alleviate sparse-reward credit assignment and reduce stage-wise optimization interference. 

% We examine whether fine-grained process rewards can stabilize RL training for our multi-step Planner--Executor--Reflector workflow by comparing MAPO with standard GRPO. Figure~\ref{fig:comp_grpo} shows that MAPO drives a faster and smoother decrease in critic loss, and reaches higher trajectory returns earlier on both the training and validation splits, while GRPO improves more slowly with larger fluctuations. At the same time, the Actor Gradient Norm exhibits lower oscillations, indicating that MAPO achieves better convergence. These observations align with our design choice to supervise intermediate trend narration, LLM decision accuracy, and reflection-based correction, and to optimize them with module-aware, group-relative advantages to alleviate sparse-reward credit assignment and reduce stage-wise optimization interference.

We examine whether fine-grained process rewards can stabilize RL training for our multi-step agentic workflow by comparing MAPO with standard GRPO. Figure~\ref{fig:comp_grpo} shows that MAPO reaches higher trajectory returns earlier on both the training and validation splits, while GRPO improves more slowly with larger fluctuations. The actor gradient norm under MAPO also exhibits smaller oscillations and a smoother overall trajectory, indicating more stable and reliable optimization dynamics throughout training. These observations align with our design choice to supervise intermediate trend narration, LLM decision accuracy, and reflection-based correction, and to optimize them with module-aware, group-relative advantages to alleviate sparse-reward credit assignment and reduce stage-wise optimization interference.

\begin{table}[t]
  \centering
  \caption{Performance consistency analysis (Weighted F1) across different lightweight predictors. Applying the GeoMind framework consistently improves performance across various backbones and datasets.}
  \label{tab:backbone_generalization}
  \renewcommand{\arraystretch}{1.15}
  \resizebox{\columnwidth}{!}{
    \begin{tabular}{lcccccc}
    \toprule
    \multirow{2}{*}{\textbf{Dataset}} & \multicolumn{2}{c}{\textbf{XGBoost}} & \multicolumn{2}{c}{\textbf{InceptionTime}} & \multicolumn{2}{c}{\textbf{GPT4TS}} \\
    \cmidrule(lr){2-3} \cmidrule(lr){4-5} \cmidrule(lr){6-7}
          & Original & w/ GeoMind & Original & w/ GeoMind & Original & w/ GeoMind \\
    \midrule
    SEAM  & 0.8355 & \textbf{0.8617} & 0.8537 & \textbf{0.8694} & 0.8425 & \textbf{0.8821} \\
    Facies & 0.4086 & \textbf{0.4469} & 0.4062 & \textbf{0.4397} & 0.4155 & \textbf{0.4582} \\
    FORCE & 0.3612 & \textbf{0.4179} & 0.3614 & \textbf{0.4056} & 0.3453 & \textbf{0.3741} \\
    GeoLink & 0.3644 & \textbf{0.3937} & 0.4116 & \textbf{0.4318} & 0.4130 & \textbf{0.4401} \\
    \bottomrule
    \end{tabular}%
  }
\end{table}

% % (2) Reflection：多源冲突下的选择偏好分析
% \subsection{Analysis of Arbitration Preferences}

% % (5) 感知岩性粒度转移：否定错误、修正结果  case study
% \subsection{Granularity Shift Detection and Correction}
% Case Study. TODO

\section{Conclusion}
This work proposed GeoMind, a process-supervised agentic workflow for lithology classification, designed to address challenges in well-log data interpretation, such as noisy signals and stratigraphic inconsistencies. By integrating specialized numerical predictors and large language model-based reasoning, GeoMind improved classification accuracy while maintaining transparent decision-making processes. The experiments demonstrated that GeoMind outperformed state-of-the-art methods across several benchmark datasets, providing consistent gains, especially in complex geological settings. Furthermore, the incorporation of process rewards and a structured Planner–Executor–Reflector workflow contributed to the stability of the framework, making it effective in handling noisy data and ambiguous boundaries. GeoMind's ability to combine fine-grained intermediate supervision with final predictions highlights its potential as a reliable tool for geoscience data analysis.

\clearpage

\bibliographystyle{ACM-Reference-Format}
\bibliography{ref}

\clearpage
\appendix

\section{Tools Description} \label{app:executor_tools}
This appendix provides a self-contained description of the tool modules used in GeoMind. For an input well-log window $X_{t:t+k-1}$, the executor invokes a set of tools to produce intermediate evidence, candidate predictions, and consistency signals. We group the tools into \emph{perception}, \emph{reasoning}, and \emph{analysis} categories, and detail their inputs/outputs and key computations below.

\noindent\textbf{Perception tools.}
\textit{Case retriever} retrieves a neighborhood of similar historical windows and their labels:
\begin{equation}
\mathcal{N}_t = \{(X^{(j)}, y^{(j)}, s_{t,j})\}_{j=1}^{K},
\label{eq:neighbors}
\end{equation}
where $s_{t,j}$ is a normalized similarity score computed by combining multiple distance metrics (Euclidean, Manhattan, and cosine) with metric-specific weights. The retrieved neighbors provide empirical evidence and local priors for downstream inference.
\textit{Trend narrator} converts multivariate log curves into a structured natural-language trend description $z_t$. The narrative summarizes depth-aware quantitative cues such as stable segments, turning points, and magnitude changes, and serves as a semantic bridge between numerical signals and LLM-based reasoning.

\noindent\textbf{Reasoning tools.}
\textit{Neighbor vote aggregator} converts the retrieved neighborhood into class confidences under a local smoothness assumption. Let $w_{t,j}\ge 0$ denote derived weights normalized as $\sum_{j=1}^{K} w_{t,j}=1$. The similarity-weighted voting confidence is
\begin{equation}
p^{\mathrm{nbr}}_{t}(c) = \sum_{j=1}^{K} w_{t,j}\,\mathbb{I}\!\left[y^{(j)}=c\right],\quad c\in\mathcal{C}.
\label{eq:knn_vote}
\end{equation}
\textit{Neural probability interpreter} takes the neural classifier softmax output $p^{\mathrm{nn}}_t\in[0,1]^{|\mathcal{C}|}$ and produces a concise natural-language explanation $e_t$ (optionally with confidence thresholding). This module improves transparency by summarizing which classes are supported by the neural predictor and how confident the model is.

\textit{Semantic reasoning engine} integrates the well-log table, the trend narrative $z_t$, and neighbor evidence $(\mathcal{N}_t, p^{\mathrm{nbr}}_t)$ to generate (i) a stepwise reasoning trace $h_t$ and (ii) an LLM candidate prediction distribution or label $\tilde{y}^{\mathrm{llm}}_t$.

\noindent\textbf{Analysis tools.}
\textit{Consensus conflict scanner} compares three predictors, namely neighbor voting, the neural classifier, and semantic reasoning. Let $\tilde{y}^{\mathrm{nbr}}_t=\arg\max_{c} p^{\mathrm{nbr}}_t(c)$ and $\tilde{y}^{\mathrm{nn}}_t=\arg\max_{c} p^{\mathrm{nn}}_t(c)$. The agreement count is defined as
\begin{equation}
A_t = \mathbb{I}[\tilde{y}^{\mathrm{nbr}}_t=\tilde{y}^{\mathrm{nn}}_t]
     +\mathbb{I}[\tilde{y}^{\mathrm{nbr}}_t=\tilde{y}^{\mathrm{llm}}_t]
     +\mathbb{I}[\tilde{y}^{\mathrm{nn}}_t=\tilde{y}^{\mathrm{llm}}_t],
\label{eq:agreement}
\end{equation}
which indicates full consensus ($A_t=3$), majority consensus (exactly two agree), or full disagreement ($A_t=0$). The scanner outputs a conflict report $q_t$ that summarizes agreement patterns and highlights regions prone to ambiguity (e.g., near lithological boundaries or transition zones).
\textit{Stratigraphic sequence validator} evaluates plausibility using a Markov transition model over lithology labels estimated from the training set. With Laplace smoothing, the transition probabilities are
\begin{equation}
P_{a\rightarrow b} = \frac{N_{a\rightarrow b}+\lambda}{\sum_{b'\in\mathcal{C}}N_{a\rightarrow b'}+\lambda|\mathcal{C}|},
\label{eq:markov}
\end{equation}
where $N_{a\rightarrow b}$ counts adjacent transitions $a\!\rightarrow\! b$ and $\lambda>0$ is the smoothing constant. Given a predicted sequence $\hat{y}_{1:T}$, the validator flags low-probability jumps based on $P_{\hat{y}_{t-1}\rightarrow \hat{y}_t}$ and outputs a validation signal $v$ (e.g., anomalous transition indicators) that can be consumed by the reflector.

\section{Inference-Consistent Training Strategy}
\label{sec:appendix_stacking}

\subsection{The Distribution Shift Challenge}
A critical bottleneck in cascading neuro-symbolic systems (such as GeoMind) is the distribution shift between the training and inference phases of the base predictors. In a naive implementation, if the numerical predictor (e.g., GPT4TS or XGBoost) and the downstream agentic workflow (Planner-Executor-Reflector) are trained sequentially on the exact same dataset $\mathcal{D}_{train}$, the numerical predictor often exhibits over-optimistic performance due to overfitting or memorization. 

Consequently, the RL agent observes a distribution of ``easy'' errors and high-confidence correct predictions during its training phase. However, during testing, the numerical predictor inevitably faces generalization errors on unseen wells. This discrepancy leads to a \textit{policy collapse}: the agent learns to overly rely on the numerical predictor's output and fails to develop robust correction strategies (e.g., triggering the Reflector to resolve conflicts) because it rarely encounters significant disagreements or low-confidence signals during training.

\subsection{K-Fold Stacking for Unbiased RL Initialization}
To address this, we introduce a \textbf{K-Fold Stacking Strategy} for generating the observation space for the RL agent. The global split is first fixed using disjoint wells; K-fold stacking is then performed exclusively within the training wells, with folds defined by well identity rather than by samples or windows. This approach allows us to construct a training set for the agent that simulates the \textit{generalization error distribution} of the base model without reducing the total volume of available training data.

% To address this, we introduce a \textbf{K-Fold Stacking Strategy} for generating the observation space for the RL agent. This approach allows us to construct a training set for the agent that simulates the \textit{generalization error distribution} of the base model without reducing the total volume of available training data.

The process follows a ``Leave-One-Fold-Out'' logic:
\begin{enumerate}
    \item We partition the training wells $\mathcal{W}_{train}$ into $K$ disjoint subsets (folds).
    \item For each fold $k$, we train a temporary base predictor $f_{\theta}^{(k)}$ on all data \textit{except} fold $k$.
    \item We then generate ``out-of-sample'' predictions (logits and probabilities) for the held-out fold $k$ using $f_{\theta}^{(k)}$.
    \item These unbiased predictions are aggregated to form the input state space for training the GeoMind agent.
\end{enumerate}

For each fold, all data-dependent state components are rebuilt from the training folds: the labeled Case Retriever excludes held-out wells, and Markov counts use only remaining labels. Held-out labels are accessed only after state construction. Thus, neither overlapping windows nor label-derived tools transfer information across folds, and $p_t^{nn}$ comes from a predictor that has not seen the held-out well. During final validation or test inference, all such components use only the corresponding permitted training wells.

% For each fold, all data-dependent components used to construct the RL state are rebuilt using only the training folds. The held-out fold labels are accessed only after state construction to compute the training signal. This strategy ensures that for every sample in the RL training set, the input features (specifically the neural probability interpreter's confidence $p_t^{nn}$) come from a model that has never seen that well's ground truth. This forces the Planner and Reflector to learn under realistic uncertainty conditions.
% During the final inference phase on the test set, we discard the temporary fold models and utilize a final base predictor $f_{\theta}^{final}$ trained on the entire training corpus $\mathcal{W}_{train}$ to maximize feature extraction capability.

\subsection{Algorithmic Formulation}
The complete training pipeline, ensuring strict separation of memorization and generalization signals, is formalized in Algorithm~\ref{alg:stacking}.

\begin{algorithm}[H]
\caption{GeoMind Robust Two-Stage Training via K-Fold Stacking}
\label{alg:stacking}
\small % Reduce font size to fit ACM column width
\begin{algorithmic}[1.2]
\REQUIRE Set of training wells $\mathcal{W} = \{W_1, \dots, W_N\}$, Base Model Architecture $\mathcal{A}$, Agent Policy $\pi_\phi$.
\STATE \textbf{Hyperparameters:} Number of folds $K$ (e.g., $K=5$).
\STATE Partition $\mathcal{W}$ into $K$ disjoint subsets: $\{\mathcal{S}_1, \dots, \mathcal{S}_K\}$.
\STATE Initialize empty container for RL observations: $\mathcal{O}_{RL} \leftarrow \emptyset$.

\STATE \textit{// Phase 1: Generate Out-of-Fold (OOF) Signals}
\FOR{$k = 1$ to $K$}
    \STATE Define training subset: $\mathcal{D}_{train}^{(k)} \leftarrow \mathcal{W} \setminus \mathcal{S}_k$
    \STATE Define validation subset: $\mathcal{D}_{val}^{(k)} \leftarrow \mathcal{S}_k$
    \STATE Train temporary predictor:
    \STATE \quad $f_{\theta_k} \leftarrow \text{Train}(\mathcal{A}, \mathcal{D}_{train}^{(k)})$
    \FOR{each well $X \in \mathcal{D}_{val}^{(k)}$}
        \STATE Generate probabilities: $\hat{p} \leftarrow f_{\theta_k}(X)$
        \STATE Construct agent state: 
        \STATE \quad $s \leftarrow \text{ConstructState}(\hat{p}, \text{Context}(X))$
        \STATE Store pair: $\mathcal{O}_{RL} \leftarrow \mathcal{O}_{RL} \cup \{(s, \text{Label}(X))\}$
    \ENDFOR
\ENDFOR

\STATE \textit{// Phase 2: Train Agent Policy}
\STATE Train Agent $\pi_\phi$ using $\mathcal{O}_{RL}$ via MAPO (optimizing for correction robustness).

\STATE \textit{// Phase 3: Finalize Base Model for Inference}
\STATE Train Master Predictor $f_{\theta^*} \leftarrow \text{Train}(\mathcal{A}, \mathcal{W})$ on \textit{entire} dataset.

\RETURN Optimized Policy $\pi_\phi$, Master Predictor $f_{\theta^*}$
\end{algorithmic}
\end{algorithm}

\section{Sample Efficiency and Convergence}
\label{sec:efficiency_analysis}

In this section, we provide a theoretical and empirical analysis of the efficiency advantages of MAPO compared with standard GRPO.

Standard reinforcement learning on agentic workflows faces a severe credit-assignment problem. In standard GRPO, the policy $\pi_\theta$ receives a reward signal $R$ only after the entire trajectory $\tau=(o_1,\dots,o_L)$ is completed. The gradient estimator typically takes the form $\nabla J \propto \sum_t \nabla \log \pi(o_t)\cdot(R-b)$. Because $R$ depends on the interactions of multiple downstream modules, it acts as a coarse and noisy proxy for the quality of early actions, such as the Planner's decision. Successful intermediate decisions may be weakened by a later error, while an ineffective decision may receive positive credit when another module recovers the trajectory. This ambiguity increases interference among module updates and requires more samples to discover consistently effective behaviors.

MAPO addresses this by introducing dense, process-level supervision. The global optimization objective is decomposed into a set of module-local objectives:
\begin{equation}
    \mathcal{J}_{MAPO}(\theta) = \sum_{m \in \mathcal{M}} \mathbb{E}\left[ \frac{1}{G} \sum_{g=1}^G \widetilde{A}_m^{(g)} \log \pi_\theta(o_m^{(g)} | q_m^{(g)}) \right]
\end{equation}
where $\widetilde{A}_m^{(g)}$ is derived from the local reward $r_m$ specific to module $m$. This formulation preserves the functional structure of the workflow during optimization: the Trend Narrator is rewarded for capturing depth-wise patterns, the Reasoning module for integrating heterogeneous evidence, and the Reflector for resolving conflicts and correcting unreliable predictions. Each module therefore receives a learning signal that directly reflects the behavior it is responsible for improving during policy optimization.

\paragraph{State Control.}
Module-level trajectories can still reach different downstream states after earlier outputs diverge. MAPO therefore scales the relative advantage within each group using the dispersion of its module states. Groups with closely aligned states retain their full update strength, whereas groups containing highly divergent contexts receive a smaller policy update. The resulting $\widetilde{A}_m^{(g)}$ combines fine-grained process supervision with state-aware attenuation, allowing strongly comparable actions to drive learning while preventing heterogeneous downstream contexts from dominating the update. This design strengthens module-local credit assignment without discarding the exploration diversity of group-relative optimization throughout the optimization process.

\paragraph{Optimization Implications.}
Together, process rewards and state-aware attenuation provide MAPO with a more targeted optimization signal than a single trajectory-level reward. The agent can reinforce effective tool selection, evidence integration, and correction behaviors independently, rather than waiting for repeated end-to-end successes before identifying useful intermediate decisions. The more direct feedback also allows correction-oriented modules to improve even when upstream predictions are imperfect, which is particularly important for GeoMind because disagreement resolution is a primary source of robustness.

\paragraph{Empirical Evidence.}
In Figure \ref{fig:comp_grpo}, this improved credit assignment translates to superior sample efficiency. Specifically, Figure \ref{fig:comp_grpo}(b) and \ref{fig:comp_grpo}(c) demonstrate that MAPO reaches a high trajectory return significantly earlier than GRPO; for instance, on the validation set, MAPO achieves a return of 0.35 at approximately 100 steps, whereas GRPO requires over 250 steps to reach the same level. Furthermore, Figure \ref{fig:comp_grpo}(a) illustrates that standard GRPO exhibits high-frequency oscillations and a generally higher gradient norm. In contrast, MAPO maintains a lower and smoother gradient norm, showing that its module-aware advantages provide a cleaner and more stable learning signal.

\section{Data Preparation and Setup}
\subsection{Dataset Specifications and Input Features}
\label{app:dataset_specs}

To facilitate reproducibility and clarify the inputs used by the GeoMind framework, we provide the detailed configuration for each benchmark dataset. Table \ref{tab:dataset_features} summarizes the specific geophysical logs and features selected as inputs for the model, along with the target lithological classes.

\begin{table}[h]
    \centering
    \caption{Detailed specification of input features and target classes for the four benchmark datasets. The input features represent the exact channels used by the GeoMind.}
    \label{tab:dataset_features}
    \small 
    \renewcommand{\arraystretch}{1}
    \renewcommand{\tabularxcolumn}[1]{m{#1}}
    \begin{tabularx}{\linewidth}{l c >{\raggedright\arraybackslash}X l}
        \toprule
        \textbf{Dataset} & \textbf{\# Cls} & \textbf{Input Features (Curves)} & \textbf{Target} \\
        \midrule
        \textbf{SEAM} & 7 & \textbf{9 Channels:} Depth, Bed Dip ($X, Y$), Total Porosity, Horizontal Resistivity, TTI Dip ($X, Y$), P-wave Velocity ($V_p$), S-wave Velocity ($V_s$) & Layers \\
        \midrule
        \textbf{Facies} & 9 & \textbf{7 Channels:} Depth, Gamma Ray (GR), Deep Induction Resistivity ($\log_{10}$), Neutron-Density Porosity Diff ($\Delta\phi$), Neutron-Density Porosity (PHIND), Photoelectric Effect (PE), Nonmarine-Marine Indicator & Lithofacies \\
        \midrule
        \textbf{FORCE} & 5 & \textbf{9 Channels:} Depth, Caliper (CALI), Sonic Slowness (DTC), Gamma Ray (GR), Neutron Porosity (NPHI), Bulk Density (RHOB), Coordinates ($X, Y, Z$) & Lithology \\
        \midrule
        \textbf{GeoLink} & 11 & \textbf{9 Channels:} Depth, Caliper (CALI), Neutron Porosity (NPHI), Bulk Density (RHOB), Gamma Ray (GR), Sonic Slowness (DTC), Resistivity (Deep, Shallow, Medium) & Lithology \\
        \bottomrule
    \end{tabularx}
    \vspace{-0.1in}
\end{table}

For the \textbf{Facies} dataset, the 9 classes correspond to specific depositional environments ranging from \textit{Nonmarine sandstone} to \textit{Phylloid-algal bafflestone}. The \textbf{FORCE} dataset targets 5 primary lithologies: \textit{Shale, Sandstone, Limestone, Marl,} and \textit{Sandstone/Shale}. \textbf{GeoLink} includes 11 granular classes, distinguishing between variations such as \textit{Silty Sand}, \textit{Cross Bedded Sand}, and \textit{Argillaceous Limestone}. Finally, \textbf{SEAM} classifies geological ages and salt bodies, including \textit{Mother Salt}, \textit{Cretaceous}, and \textit{Lower Miocene} layers.

\subsection{Datasets Description}
\label{dataset_more}
We evaluate our method on four public well-log datasets collected from established open repositories for comprehensive evaluation. Specifically, we use the following datasets:

\begin{itemize}[leftmargin=*]
 \item \textbf{SEAM\footnote{\url{https://wiki.seg.org/wiki/Open_data\#Well_logs}}}: the SEG Wiki Open Data "Well logs" catalog as an entry point to publicly released well-log resources for reproducible geoscience benchmarking;

 \item \textbf{Facies\footnote{\url{https://www.kaggle.com/datasets/imeintanis/well-log-facies-dataset}}}: the Kaggle Well Log Facies Dataset, which contains facies logs from 7 wells in the Council Grove gas reservoir (Kansas) with facies labels derived from core observations;

 \item \textbf{Force\footnote{\url{https://www.kaggle.com/datasets/faresazzam/well-logs-dataset-for-machine-learning}}}: the Kaggle Well logs dataset for machine learning, released for lithology (rock-type) prediction from multiple geophysical well-log measurements and associated with the FORCE 2020 lithology prediction context;

 \item \textbf{Geolink\footnote{\url{https://github.com/LukasMosser/geolink_dataset?tab=readme-ov-file}}}: the GEOLINK-S2 well-log dataset accessed via the geolink\_dataset repository, which provides analysis notebooks and preprocessing code for the GEOLINK-S2 data.
\end{itemize}

\subsection{Data Preprocessing and Windowing}
\label{app:preprocessing}
To ensure high-quality input signals for the executor tools, we applied a standardized preprocessing pipeline across all datasets prior to tokenization or numerical embedding.

\paragraph{Outlier Handling and Imputation.} Raw well-log measurements often contain sensor errors or null values due to borehole washout or instrument failure. We first removed physical outliers based on domain knowledge (e.g., removing negative values for Gamma Ray or Density). Missing intervals shorter than 2 meters were filled using linear depth-wise interpolation, while sequences with larger gaps were discarded to maintain continuity.

\paragraph{Feature Normalization.} Given the diverse physical units of the logs, we applied channel-specific normalization:
\begin{itemize}
    \item \textbf{Logarithmic Transformation:} For resistivity logs (e.g., \texttt{RDEP}, \texttt{RMED}), which span multiple orders of magnitude, we applied a $\log_{10}$ transformation to compress the dynamic range: $x' = \log_{10}(x + \epsilon)$.
    \item \textbf{Z-Score Standardization:} All other linear curves (e.g., \texttt{GR}, \texttt{NPHI}, \texttt{RHOB}) were standardized to zero mean and unit variance: $x' = (x - \mu) / \sigma$, where statistics are computed globally on the training set.
\end{itemize}

% \paragraph{Sliding Window Generation.} To capture local stratigraphic context, we sliced the depth-aligned sequences into fixed-length windows. We utilized a window size of $L=16$ depth samples with a stride of $S=4$ for training, and non-overlapping windows for testing. This window length was empirically selected to include sufficient context for the \textit{Trend Narrator} to identify meaningful geological patterns (e.g., upward-fining sequences) without introducing excessive noise.
\paragraph{Sliding Window Generation.} Complete wells were assigned to disjoint train/validation/test partitions before window generation. Training used length-$16$ windows with stride $4$, whereas validation and testing used non-overlapping windows ($S=L=16$); overlap therefore cannot cross partitions or K-fold held-out wells. This window length was empirically selected to include sufficient context for the \textit{Trend Narrator} to identify meaningful geological patterns (e.g., upward-fining sequences) without introducing excessive noise.

\begin{table*}[h]
\centering
% \small
\renewcommand{\arraystretch}{1.18}
\setlength{\tabcolsep}{7pt}
\caption{End-to-end inference efficiency of GeoMind. The statistics are averaged over test wells.}
\label{tab:efficiency}
\begin{tabular}{lrrrrr}
\toprule
Dataset
& Windows / well
& Tool calls / well
& Calls / window
& Time / window (s)
& Time / well (min) \\
\midrule
SEAM    & 89  & 401       & 4.5 & 1.42 & 2.1 \\
Facies  & 28  & 123       & 4.4 & 1.71 & 0.8 \\
FORCE   & 300 & $1{,}410$ & 4.7 & 1.32 & 6.6 \\
GeoLink & 283 & $1{,}329$ & 4.7 & 1.31 & 6.2 \\
\bottomrule
\end{tabular}
\end{table*}

\section{Implementation and Reproducibility Details}
\label{imp_appendix}

To ensure the reproducibility of GeoMind, we provide the implementation details for our agentic workflow.

\noindent\textbf{Stage 1: Domain-Adaptive SFT.}
We fine-tune the base model using the curated instruction dataset (described in Sec. \ref{sec:sft}) to adapt it to geological semantics. Specifically, we employ the AdamW optimizer with $\beta_1=0.9$ and $\beta_2=0.95$. The training process is conducted with a global batch size of 64 over 3 epochs, incorporating a linear warmup phase of 0.03 epochs. We initialize the learning rate at $2\times 10^{-5}$ and apply a cosine decay scheduler reducing it to $2\times 10^{-6}$. The maximum sequence length is set to 8192 tokens to accommodate long-context geological reasoning.

\noindent\textbf{Stage 2: MAPO Training.}
% For the reinforcement learning phase, we employ Module-Aware GRPO to align the agent with process rewards. We sample $G=8$ trajectories per input for group-relative advantage estimation and set the KL coefficient $\beta$ to $0.04$ to prevent excessive deviation from the SFT reference model. We use a single state-gate coefficient $\gamma=1.0$ across all modules. The learning rate is reduced to $5\times 10^{-6}$ to ensure stable policy updates. 
For the reinforcement learning phase, we employ Module-Aware GRPO to align the agent with process rewards. We sample $G=8$ trajectories per input for group-relative advantage estimation and set the KL coefficient $\beta$ to $0.04$ to prevent excessive deviation from the SFT reference model. Prompt embeddings are obtained by masked mean pooling the last-layer prompt representations of the frozen reference model, followed by $\ell_2$ normalization. We reuse the existing reference-model pass and stop gradients through these embeddings, adding no extra model call. We use a single coefficient $\gamma=0.6$ across all modules. The learning rate is reduced to $5\times 10^{-6}$ to ensure stable policy updates. 
% Furthermore, to encourage exploration and avoid premature convergence, we adopt the asymmetric clipping strategy from DAPO \cite{yu2025dapo} with $\epsilon_{low}=0.2$ and $\epsilon_{high}=0.28$. This configuration allows the agent to retain updates that yield significant advantage improvements while strictly bounding those that might degrade stability.

\section{Additional Controlled Analyses}
\label{app:additional_controlled_analyses}

\begin{table}[t]
\centering
\small
\renewcommand{\arraystretch}{1.2}
\setlength{\tabcolsep}{4.2pt}
\caption{Controlled attribution analyses across four benchmarks. We report F1 scores.}
\vspace{-0.08in}
\begin{tabular}{
    @{} l
    S[table-format=1.4]
    S[table-format=1.4]
    S[table-format=1.4]
    S[table-format=1.4]
    @{}
}
\toprule
\textbf{Method} 
& {\textbf{SEAM}} 
& {\textbf{Facies}} 
& {\textbf{FORCE}} 
& {\textbf{GeoLink}} \\
\midrule
\textbf{GeoMind (MAPO)}
& \textbf{0.8821} & \textbf{0.4582} & \textbf{0.3741} & \textbf{0.4401} \\
\midrule
\multicolumn{5}{@{}l}{\textit{Static evidence aggregation}} \\
Majority Voting
& 0.8708 & 0.4463 & 0.3624 & 0.4283 \\
Fixed Fusion
& 0.8747 & 0.4496 & 0.3651 & 0.4327 \\
\midrule
\multicolumn{5}{@{}l}{\textit{Process-reward optimization}} \\
MAPO w/o State Gate
& 0.8786 & 0.4527 & 0.3693 & 0.4348 \\
PPO w/ Process Rewards
& 0.8673 & 0.4386 & 0.3564 & 0.4218 \\
\bottomrule
\end{tabular}
\vspace{-0.14in}
\label{tab:controlled_attribution}
\end{table}

This appendix provides three additional controlled analyses to further clarify the source of GeoMind's improvements under controlled experimental settings. 
These experiments are complementary to the main ablation study. 
Rather than exhaustively removing every module, we focus on three specific questions raised by the ablation analysis: 
whether GeoMind is merely an ensemble of multiple predictors, whether the process rewards alone can explain the training gain, and how much the stratigraphic sequence validator contributes to the final performance.

\subsection{Static Aggregation vs. Agentic Control}
\label{app:static_aggregation}

Given that GeoMind integrates multiple evidence sources, including neural predictions, neighbor-based evidence, LLM reasoning, and geological validation signals, it is important to distinguish the effect of static evidence aggregation from that of agentic control. To this end, we compare GeoMind with two non-agentic aggregation variants: a majority-vote variant that aggregates candidate predictions without reflection-based arbitration, and a fixed-fusion variant that uses the same evidence sources but replaces adaptive planning and reflection with a fixed aggregation rule. As shown in Table~\ref{tab:controlled_attribution}, both variants underperform the GeoMind framework. Although fixed fusion performs better than majority voting, it consistently lags behind GeoMind, suggesting that the performance gain comes not only from combining several predictors but also from adaptive tool orchestration and reflection-based conflict resolution. This controlled comparison helps isolate the contribution of dynamic decision-making under otherwise matched evidence access. The results further indicate that GeoMind benefits from how evidence is selected, reconciled, and revised, rather than merely from the availability of additional signals.

\subsection{Process-Reward Optimization Analysis}
\label{app:process_reward_optimization}

% We further examine whether the gain of MAPO can be fully explained by dense process rewards, or whether the optimization strategy used to exploit these rewards also matters. To this end, we compare GeoMind with a PPO-based variant that uses the same process rewards. This variant preserves intermediate supervision but removes MAPO's module-aware group-relative credit assignment. As shown in Table~\ref{tab:controlled_attribution}, the PPO alternative underperforms GeoMind, indicating that dense rewards alone are not sufficient to match the full training strategy. The results suggest that GeoMind benefits not only from reward design, but also from assigning and optimizing these rewards at the module level during multi-step workflow optimization. In addition, PPO typically relies on an additional value function for advantage estimation, introducing extra memory and optimization overhead compared with the value-free group-relative update used in MAPO.

We further examine whether the gain of MAPO can be fully explained by dense process rewards, or whether the optimization strategy used to exploit these rewards also matters. To this end, we compare GeoMind with a PPO-based variant that uses the same process rewards. This variant preserves intermediate supervision but removes MAPO's module-aware group-relative credit assignment.
We also include MAPO w/o State Gate, which removes the state-dispersion attenuation and directly applies the unattenuated module-level group-relative advantage. As shown in Table~\ref{tab:controlled_attribution}, removing the state gate consistently degrades performance across all four benchmarks. Although the reductions are moderate, their consistency indicates that state-aware attenuation improves optimization by suppressing updates driven by heterogeneous downstream states. Together with the larger performance gap between MAPO and PPO with process rewards, these results distinguish the complementary contributions of dense process supervision, module-level group-relative credit assignment, and state-aware stabilization.

\subsection{Stratigraphic Sequence Validator Analysis}
\label{app:validator_analysis}

\begin{table}[t]
\centering
\small
\renewcommand{\arraystretch}{1.2}
\setlength{\tabcolsep}{4.2pt}
\caption{Stratigraphic validator analysis across four benchmarks. We report F1 scores.}
\vspace{-0.08in}
\begin{tabular}{
    @{} l
    S[table-format=1.4]
    S[table-format=1.4]
    S[table-format=1.4]
    S[table-format=1.4]
    @{}
}
\toprule
\textbf{Method} 
& {\textbf{SEAM}} 
& {\textbf{Facies}} 
& {\textbf{FORCE}} 
& {\textbf{GeoLink}} \\
\midrule
\textbf{GeoMind (Ours)}
& \textbf{0.8821} & \textbf{0.4582} & \textbf{0.3741} & \textbf{0.4401} \\
\midrule
w/o Stratigraphic Validator
& 0.8732 & 0.4449 & 0.3617 & 0.4286 \\
w/ Uniform Transition Prior
& 0.8691 & 0.4398 & 0.3572 & 0.4241 \\
w/ Random Transition Prior
& 0.8647 & 0.4351 & 0.3536 & 0.4194 \\
\bottomrule
\end{tabular}
\vspace{-0.1in}
\label{tab:validator_analysis}
\end{table}

Table~\ref{tab:validator_analysis} further analyzes the role of the Stratigraphic Sequence Validator. 
Removing the validator consistently reduces performance across all four benchmarks, showing that geological transition information is useful. 
Replacing the learned transition matrix with a uniform or random transition prior leads to further degradation, indicating that the benefit does not come from generic smoothing, but from meaningful transition statistics estimated from the training labels. 
At the same time, the moderate gap between GeoMind and the variant without the validator suggests that the validator is helpful but not the sole source of improvement.

\section{Inference Efficiency and Practical Trade-offs}
\label{app:efficiency}

\begin{figure*}[ht]
\centering
\includegraphics[width=0.8\textwidth]{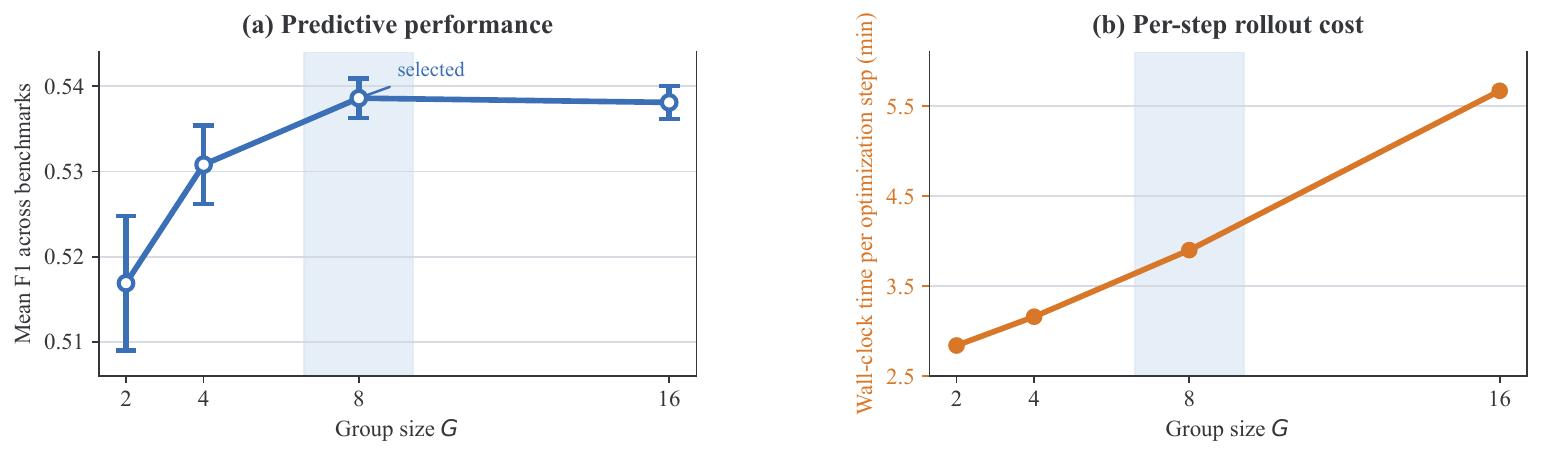}
\vspace{-0.06in}
\caption{Sensitivity of MAPO to the group size $G$. Panel (a) reports the mean F1 across the four benchmarks, with error bars indicating variability across runs. Panel (b) reports the rollout time per optimization step. The shaded region marks the default setting $G=8$ to highlight its balance between performance and efficiency.}
\vspace{-0.08in}
\label{fig:rollout_n_sensitivity}
\end{figure*}

Table~\ref{tab:efficiency} summarizes the end-to-end inference cost of GeoMind. The workflow requires approximately 4.4--4.7 tool calls and three LLM forward passes per window, resulting in 1.31--1.71 seconds of latency per window. The total runtime mainly depends on the number of windows in each well, ranging from 0.8 minutes on Facies to 6.6 minutes on FORCE. Although this is slower than neural-only baselines, not all tool calls involve generative inference; retrieval, voting, conflict detection, and transition validation are lightweight operations. Moreover, the Planner adaptively selects tools instead of executing a fixed maximal pipeline for every input.

Lithology interpretation provides a fundamental basis for characterizing subsurface geological structures and guiding subsequent mineral exploration and resource development. Inaccurate lithology classifications may propagate through downstream workflows, distort reservoir or resource assessments, and lead to inappropriate drilling, extraction, or development decisions, thereby incurring substantial operational and economic costs. In this offline and high-stakes setting, predictive accuracy and geological consistency are therefore considerably more important than maximizing inference throughput. From this perspective, the minutes-scale runtime required by GeoMind for each well represents a reasonable computational trade-off for obtaining more reliable, geologically coherent, and verifiable interpretations.

\section{Group-Size Sensitivity of MAPO}

% Figure~\ref{fig:rollout_n_sensitivity} summarizes the sensitivity of MAPO to $n\in \{2,4,8,16\}$. With $n=2$, group-relative normalization relies on too few trajectories to form a reliable baseline, leading to noisier advantage estimates and larger run-to-run variation. Increasing the group size progressively improves predictive performance and optimization stability, while gains become small once $n$ reaches 8. Further increasing the group size to $n=16$ yields negligible improvement over $n=8$, but substantially increases rollout time and peak memory. We therefore use $n=8$ as the default because it provides the best balance between stable group-relative estimation, predictive performance, and computational overhead.
Figure~\ref{fig:rollout_n_sensitivity} summarizes the sensitivity of MAPO to $G\in \{2,4,8,16\}$. With $G=2$, group-relative normalization relies on too few trajectories to form a reliable baseline, leading to noisier advantage estimates and larger run-to-run variation. Increasing the group size progressively improves predictive performance and optimization stability, while gains become small once $G$ reaches 8. Further increasing the group size to $G=16$ yields negligible improvement over $G=8$, but substantially increases rollout time and peak memory. We therefore use $G=8$ as the default because it provides the best balance between stable group-relative estimation, predictive performance, and computational overhead.

\section{Leave-One-Region-Out Transfer Evaluation}
\label{app:leave_one_region}

To construct a stricter geological-transfer test, we partition the FORCE wells into three spatial regions by applying K-Means to their well coordinates. Each fold uses two regions for training and treats the remaining region as unseen test data. All data-dependent components, including GPT4TS, the retrieval index, the stratigraphic transition model, and the SFT and MAPO stages, are fitted using only the two training regions; labels from the held-out region are excluded from training and model selection.

\begin{table}[tbh]
\centering
\small
\renewcommand{\arraystretch}{1.2}
\setlength{\tabcolsep}{6.5pt}
\caption{Leave-one-region-out transfer evaluation on FORCE. We report F1 scores; Gain denotes the absolute improvement of GeoMind over GPT4TS.}
\label{tab:leave_one_region}
\vspace{-0.1in}
\begin{tabular}{@{}lccc@{}}
\toprule
\textbf{Setting} & \textbf{GPT4TS} & \textbf{GeoMind} & \textbf{Gain} \\
\midrule
Standard split       & 0.3453 & \textbf{0.3741} & +0.0288 \\
Leave-one-region-out & 0.2875 & \textbf{0.3329} & +0.0454 \\
\bottomrule
\end{tabular}
\vspace{-0.02in}
\end{table}

As shown in Table~\ref{tab:leave_one_region}, the regional shift lowers the F1 of both methods, confirming that the held-out split is substantially more challenging than the standard evaluation. Nevertheless, GeoMind retains a clear advantage, and its margin over GPT4TS increases from 0.0288 to 0.0454. This result is consistent with the Planner skipping unreliable evidence and the Reflector overriding a mismatched validator signal when numerical and semantic evidence disagree. It supports improved robustness under the evaluated regional shift.

\section{Over-Segmentation Analysis}
\label{app:fragmentation_metric}

The main-text fragmentation rate identifies predicted runs that are anomalously short relative to the class-conditioned lower tail of the training ground-truth thickness distribution. As a complementary view, we directly test whether a method subdivides annotated lithological layers. Let $\mathcal{R}_d$ denote the set of maximal contiguous runs in the ground-truth label sequences of dataset $d$, with runs pooled across wells but never across well boundaries. For a ground-truth run $r=[s_r,e_r]$, we count the number of predicted label transitions occurring strictly inside that run:
\begin{equation}
n_r(\hat y)=\sum_{t=s_r}^{e_r-1}\mathbb{I}[\hat y_t\neq\hat y_{t+1}].
\end{equation}
The complementary GT-relative over-segmentation rate is
\begin{equation}
\operatorname{OverSeg}_d
=\frac{100}{|\mathcal{R}_d|}
\sum_{r\in\mathcal{R}_d}\mathbb{I}[n_r(\hat y)>0].
\label{eq:gt_relative_fragmentation}
\end{equation}
This metric is the percentage of annotated lithological layers split by at least one predicted internal transition. Unlike the main metric, it does not use a run-length threshold and does not penalize a correctly recovered thin bed solely because it is short. The same ground-truth runs and definition are used for every method, making this analysis complementary rather than a replacement for Figure~\ref{fig:frag}.

\begin{table}[th]
\centering
\footnotesize
\renewcommand{\arraystretch}{1.15}
\caption{Complementary GT-relative over-segmentation rate (\%); lower is better.}
\label{tab:gt_fragmentation}
\vspace{-0.07in}
\begin{tabular*}{\columnwidth}{@{\extracolsep{\fill}}lrrr@{}}
\toprule
\textbf{Dataset} & \textbf{XGBoost} & \textbf{GPT4TS} & \textbf{GeoMind} \\
\midrule
Facies  & 18.74 & 16.81 & \textbf{13.42} \\
FORCE   & 21.63 & 18.97 & \textbf{15.61} \\
GeoLink & 19.41 & 17.25 & \textbf{14.38} \\
\midrule
Macro Avg. & 19.94 & 17.48 & \textbf{14.47} \\
\bottomrule
\end{tabular*}
\vspace{-0.10in}
\end{table}

\clearpage

\end{document}